\pdfoutput=1 

\documentclass[journal,twoside,web]{ieeecolor}

\usepackage{generic}
\usepackage{amsmath,amssymb,amsfonts}
\usepackage{graphicx}
\usepackage{hyperref}
\hypersetup{hidelinks=true}
\usepackage{textcomp}
\usepackage{makecell}
\usepackage{threeparttable} 
\usepackage{listings}
\usepackage{longtable}

\let\labelindent\relax
\usepackage{enumitem}
\usepackage{chngcntr}
\usepackage{booktabs}
\usepackage{etoc}

\usepackage{lipsum}
\usepackage{authblk}
\usepackage[T1]{fontenc}    
\usepackage{diagbox}
\usepackage{cleveref}
\usepackage{amssymb}
\usepackage{enumitem}
\usepackage{arydshln}
\usepackage{multirow}
\usepackage{ragged2e}
\usepackage{mathtools}
\usepackage{graphicx}
\usepackage[normalem]{ulem}
\usepackage{dsfont}
\usepackage{soul}
\usepackage{tabularx}
\usepackage{algorithm}
\usepackage{algpseudocode} 

\usepackage[backend=biber, style=ieee, sorting=none]{biblatex} 
\def\BibTeX{{\rm B\kern-.05em{\sc i\kern-.025em b}\kern-.08em
    T\kern-.1667em\lower.7ex\hbox{E}\kern-.125emX}}
\begin{document}
\title{DR-net-Mamba: Selective State-Space Modeling for Long-Range ECG Time-Series Denoising}
\author{Basile Morel, Samuel Ruip\'erez-Campillo, Andreas P. Streich, Julia E. Vogt, Thomas Hofmann
\thanks{B. Morel, S. Ruip\'erez-Campillo, and A. P. Streich are co-first authors. J.E. Vogt and T. Hofmann as co-senior authors.}
\thanks{B. M., S. R.-C., A. P. S., J.E. V., and T. H. and  are with the Department of Computer Science at ETH Zurich, Universit\"atstrasse 6, 8092 Z\"urich, Switzerland.}
}

\maketitle

\begin{abstract}
Electrocardiogram (ECG) recordings are corrupted by non-stationary noise sources that degrade diagnostic reliability, particularly in ambulatory and long-duration recordings. Deep learning denoisers exist, but convolutional architectures are limited by their receptive field, transformer-based models scale quadratically with sequence length, and diffusion-based approaches incur prohibitive inference cost. We propose a Mamba-augmented model that inserts selective state-space blocks at the convolutional bottleneck, combining local feature extraction with long-range temporal modeling at linear complexity. We comprehensively evaluate the proposed model with respect to reconstruction fidelity, noise robustness, recording-length scaling, and downstream diagnostic classification across over 40 pathology classes. On synthetic and real datasets, our model achieves the highest SNR and lowest RMSE, with the Mamba advantage increasing with sequence length and in low-SNR regimes. On classification with two independent classifiers, the proposed Mamba-based models achieve the best macro AUROC among all denoisers and improve over their convolutional base models. 
Calibration is more nuanced and classifier-dependent: denoising improves Binary Cross-Entropy and Brier score on Inception1D but often fails to beat the noisy input on ResNet1D-Wang, and the lead-specific Mamba variant is the only denoiser to improve both calibration metrics over the noisy baseline on both classifiers. Per-class analysis reveals a morphology-dependent benefit: Mamba substantially improves ST/T-change diagnoses, which depend on broad, context-sensitive waveforms.
\end{abstract}


\etocdepthtag.toc{main}
\section{Introduction}

Cardiovascular diseases remain the leading cause of mortality worldwide~\cite{chong2025global}, motivating the development of scalable, automated screening tools. The electrocardiogram (ECG) is non-invasive, inexpensive, and widely deployed across clinical and ambulatory settings \cite{rowin2025extended}. Recent work has shown that deep learning (DL) applied to ECG signals can aid in the detection and risk assessment of multiple cardiovascular conditions \cite{tian2024foundation, li2025electrocardiogram}, including structural abnormalities \cite{poterucha2025detecting}, ventricular dysfunction in adult \cite{vaid2022using} and pediatric populations \cite{mayourian2024pediatric, mayourian2025electrocardiogram}, ischemic disease \cite{yu2025ecg}, cardiomyopathies \cite{sangha2025identification}, arrhythmias \cite{kolk2024dynamic, ruiperez2024clustering}, or primary prevention \cite{kolk2023optimizing}. However, 
downstream analyses depend critically on signal quality, which is often degraded by non-stationary, unstructured noise sources that evolve unpredictably over long recordings.
When multiple contaminations overlap, the signal-to-noise ratio (SNR) can drop to levels at which both human readers and automated classifiers fail.

Traditional denoising methods based on filtering, wavelets, and signal decomposition can be effective in controlled settings but require expert-driven parameter tuning and degrade under realistic multi-source noise~\cite{chatterjee2020review, jia2024preprocessing}. DL has advanced the field substantially: convolutional autoencoders~\cite{Chiang_DAE, Qiu2021ECGDenoising}, recurrent networks~\cite{antczak2019deeprecurrentneuralnetworks}, GANs~\cite{wang2022ecg, singh2021new}, and diffusion models~\cite{li2024descod} all improve reconstruction fidelity over classical pipelines. Yet, supervised methods suffer from the scarcity of clean ground truth in clinical datasets~\cite{Chiang_DAE}, Transformer-based architectures face quadratic compute and memory scaling with sequence length~\cite{vaswani2017attention, tay2022efficient}, and diffusion models incur high inference costs from iterative sampling~\cite{song2022denoisingdiffusionimplicitmodels}. These constraints are especially problematic for long ambulatory recordings, where denoising must operate over tens of seconds of continuous signal at low latency.

Selective state-space models, and Mamba in particular~\cite{gu2024mamba}, offer a compelling alternative. By making the state-transition parameters input-dependent, Mamba combines long-range temporal modeling with linear-time complexity and efficient recurrent-style inference. Mamba has been explored for ECG classification~\cite{qiang2024ecgmamba, xu2024mambacapsule} and short-segment enhancement~\cite{hung2024mecgemambabasedecgenhancer}, yet its potential for ECG denoising under diverse noise conditions and recording lengths remains unexplored.

In this work, we propose \emph{DR-net-Mamba}, a two-stage architecture that inserts selective state-space blocks at the bottleneck of a convolutional UNet encoder--decoder, extending the effective receptive field to the full sequence length while adding only linear computational overhead, with a compact architecture suited for deployment. A learnable log-compression layer normalizes dynamic range. We evaluate DR-net-Mamba across four complementary axes: (i) reconstruction fidelity on synthetic, clinical, and ambulatory datasets under simultaneous four-source noise at clinically calibrated levels; (ii) noise robustness across clinically representative SNR regimes and noise conditions; (iii) recording-length scaling, characterizing how the selective state-space bottleneck leverages increasing temporal context to improve sequence-to-sequence modeling; and (iv) downstream diagnostic classification of 44 PTB-XL pathology classes through two independent classifiers, evaluating whether denoising gains preserve or improve diagnostic accuracy in a clinically actionable setting. To the best of our knowledge, this is the first evaluation framework spanning reconstruction, noise robustness, length scaling, and clinical utility within a single study for ECG denoising. 
Our open code is publicly available at \url{https://github.com/...}. 




\section{Related Work}

\paragraph{Traditional ECG denoising.}
Classical approaches to ECG denoising rely on frequency-domain filtering, adaptive and Kalman-based methods~\cite{vullings2010adaptive}, empirical mode decomposition~\cite{blanco2008ecg}, independent component analysis~\cite{lee2022ecg}, and wavelet transforms~\cite{yu2024accurate}. These methods require expert-driven parameter selection and degrade with non-stationary artifacts~\cite{holgado2023characterization}, motivating data-driven alternatives.

\paragraph{Deep learning for ECG denoising.}
While DL has recently been applied to denoising across multiple cardiac signal domains \cite{yang2025ecgdedrdnet, avila2025generative, ruiperez2026reducing, ruiperez2026antithetic}, the ECG is the most extensively studied, with data-driven methods learning denoising priors directly from examples rather than relying on hand-crafted filter design. Convolutional architectures range from fully convolutional autoencoders~\cite{Chiang_DAE} to multi-scale UNet variants with channel attention and detail-restoration stages~\cite{Qiu2021ECGDenoising}. Recurrent approaches, including LSTM-based denoising autoencoders~\cite{antczak2019deeprecurrentneuralnetworks} and deeper recurrent--convolutional hybrids~\cite{hou2023deep}, model temporal dependencies explicitly but scale poorly to long sequences. Multi-scale filtering designs~\cite{romero2021deepfilter} and transformer-based denoisers~\cite{zhu2024ecg} further improve fidelity, though transformers incur quadratic complexity in sequence length~\cite{vaswani2017attention}. On the generative side, GAN-based frameworks~\cite{wang2022ecg, singh2021new} and score-based diffusion models~\cite{li2024descod} achieve strong reconstruction quality and can support unpaired training via cycle-consistency~\cite{kiranyaz2022blindecgrestorationoperational}, but diffusion models face high inference costs from iterative sampling~\cite{song2022denoisingdiffusionimplicitmodels}. A persistent challenge across supervised methods is the scarcity of clean ground truth in clinical datasets, where residual artifacts survive preprocessing and become part of the reconstruction target~\cite{Chiang_DAE, Qiu2021ECGDenoising}.

\paragraph{Synthetic ECG generation for evaluation.}
The absence of clean clinical references motivates the use of synthetic signals for controlled evaluation: we cannot access the true latent signal and its unknown generative process beneath observed noisy measurements. The dynamical model of \cite{mcsharry2003dynamical} generates realistic PQRST morphology from a three-variable ODE, producing signals of arbitrary length without segment stitching and requiring no training data. Data-driven generators based on GANs~\cite{delaney2019synthesis, zhu2019electrocardiogram} and VAEs~\cite{kuznetsov2021vae} can capture richer morphological distributions but inherit residual noise from their training corpora, limiting their utility as denoising benchmarks. We adopt the ODE-based generator for our synthetic experiments, as it provides verified clean ground truth and continuous-time integration.

\paragraph{Selective state-space models and Mamba.}
Structured state-space models (SSMs) offer linear-time sequence modeling as an alternative to attention~\cite{gu2024mamba}. Mamba extends SSMs with input-dependent selectivity: the state-transition parameters $(\mathbf{B}, \mathbf{C}, \Delta)$ are computed from each input token, enabling the model to selectively propagate or forget information along the sequence. This yields time-varying dynamics with efficient hardware-aware implementations and recurrent-style inference, and can be interpreted through an implicit attention viewpoint~\cite{ali2025hidden}.

Interest in Mamba for physiological signals has grown rapidly~\cite{sellam2025mamba}, and existing ECG applications fall into two groups. For \emph{classification}, ECGMamba~\cite{qiang2024ecgmamba} replaces attention with stacked selective state-space blocks to categorize rhythms and morphologies, while MambaCapsule~\cite{xu2024mambacapsule} couples a Mamba backbone with a capsule network for interpretable disease diagnosis; both target discrete labels rather than waveform reconstruction. For \emph{enhancement}, MECGE~\cite{hung2024mecgemambabasedecgenhancer} is, to our knowledge, the only prior Mamba-based ECG denoiser: it transforms the signal to the time--frequency domain via the STFT and applies bidirectional Mamba blocks along both the time and frequency axes to enhance the resulting spectrogram. This performs well on short segments, but its per-frame frequency sequences cause memory to scale with recording length, limiting use on long ambulatory data. 
Our work differs from these prior ECG applications of Mamba in three respects: (i) we operate directly in the time domain rather than on spectrograms, (ii) we use Mamba as a bottleneck augmentation to a convolutional backbone rather than as a standalone architecture, and (iii) we evaluate across reconstruction, robustness, length scaling, and downstream classification within a unified framework.

\section{Methods}\label{sec:methods}
Let $\mathbf{x} \in \mathbb{R}^{T}$ denote a clean single-lead ECG signal of $T$ time steps. The observed signal $\tilde{\mathbf{x}} = \mathbf{x} + \mathbf{v}$ is corrupted by zero-mean noise $\mathbf{v}$ comprising baseline wander, muscle artifact, electrode motion, and additive white Gaussian noise. An offline denoising model $\varphi$ produces the estimate $\hat{\mathbf{x}} = \varphi(\tilde{\mathbf{x}}) \approx \mathbf{x}$ from the full observed sequence.

\begin{figure*}[t]
    \centering
    \includegraphics[width=\linewidth]{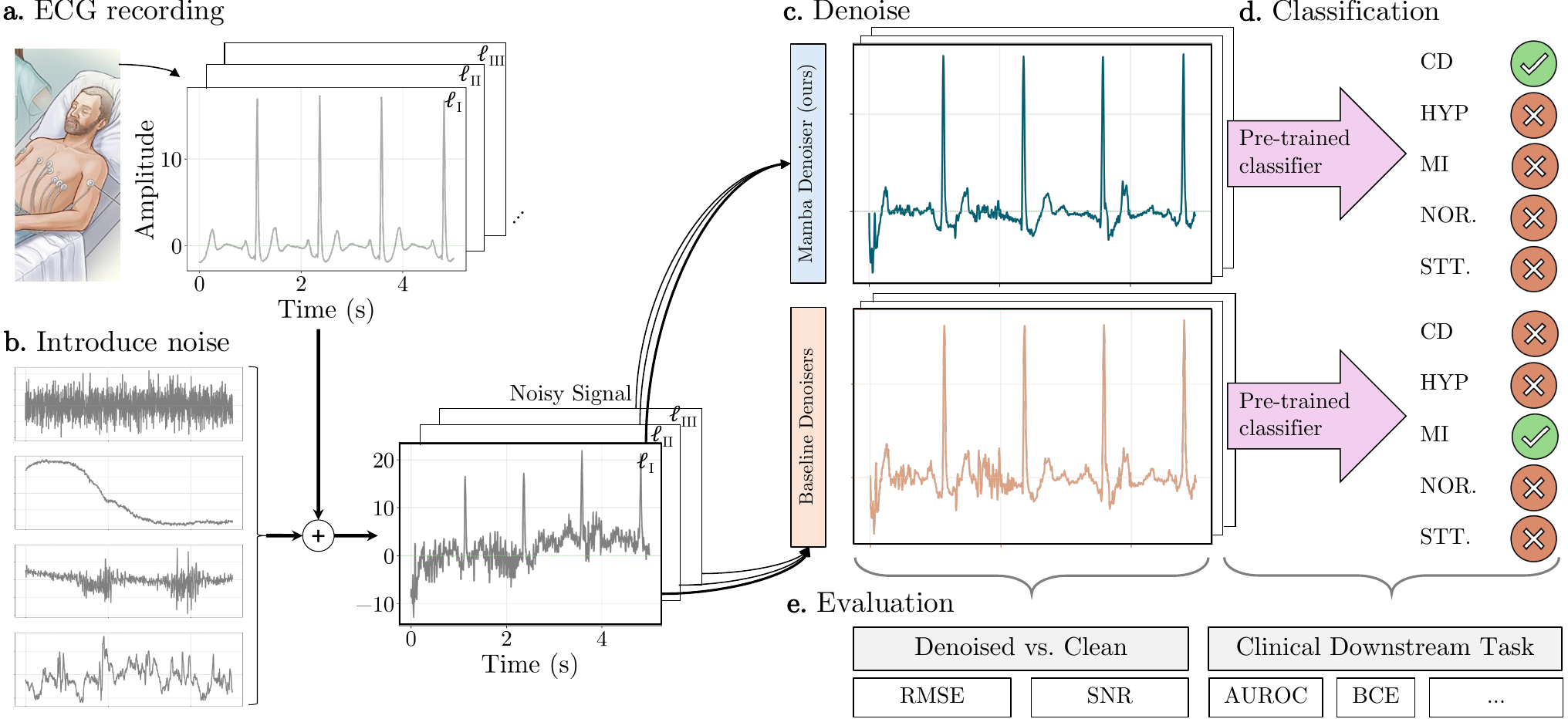}
    \caption{Overall Pipeline: clean ECG records (a) are corrupted with noise (b), denoised (c), and classified (d) to assess whether reconstruction gains translate to diagnostic accuracy (e).}
    \label{fig:ds_eval}
\end{figure*}

\subsection{Architecture}

\subsubsection{Baseline architectures.} 
We benchmark against six models spanning the major architectural families. \textbf{UNet}~\cite{Qiu2021ECGDenoising}: a three-stage convolutional encoder--decoder adapted from the original 2D U-Net~\cite{ronneberger2015unetconvolutionalnetworksbiomedical} for 1D signals, with decreasing kernel sizes to separate low- and high-frequency components. \textbf{IMUNet}~\cite{Qiu2021ECGDenoising}: extends UNet with deeper convolutional blocks, channel attention, additional skip connections, and a dilated context-contrast bottleneck. \textbf{DR-net-UNet} and \textbf{DR-net-IMUNet}~\cite{Qiu2021ECGDenoising}: two-stage pipelines that cascade either backbone with a multi-branch detail-restoration network operating at full resolution to recover fine waveform features. \textbf{DRNN}~\cite{antczak2019deeprecurrentneuralnetworks}: an LSTM-based denoising autoencoder that processes the signal sample-by-sample. \textbf{DAE}~\cite{Chiang_DAE}: a fully convolutional autoencoder with 13 symmetric layers and stride-based downsampling. \textbf{MECGE}~\cite{hung2024mecgemambabasedecgenhancer}: a Mamba-based enhancer operating in the STFT domain (described above). These baselines span convolutional, recurrent, and state-space paradigms, providing a comprehensive reference for evaluating the proposed architecture.

\subsubsection{Convolutional Backbone and its Receptive-Field Limitation}
We adopt a 1D UNet encoder--decoder~\cite{Qiu2021ECGDenoising, ronneberger2015unetconvolutionalnetworksbiomedical} as our backbone. The encoder consists of three stages with convolutional kernel sizes decreasing from 25 to 3, interleaved with pooling layers (factors $1/5$, $1/2$, $1/2$) that progressively down-sample the input by a factor of 20. Skip connections link each encoder stage to the corresponding decoder stage, and bilinear upsampling restores the original resolution. While effective at extracting local features, the convolutional receptive field is bounded: a layer-by-layer analysis (Supplementary Material~\ref{app:receptive}) yields the receptive field $r_0 = \sum_{l=1}^{L} \bigl( (k_l - 1) \prod_{i=1}^{l-1} s_i \bigr) + 1 = 542$ time steps---approximately 1.5\,s at 360\,Hz.

\begin{figure*}[t]
  \centering
  \includegraphics[width=\linewidth]{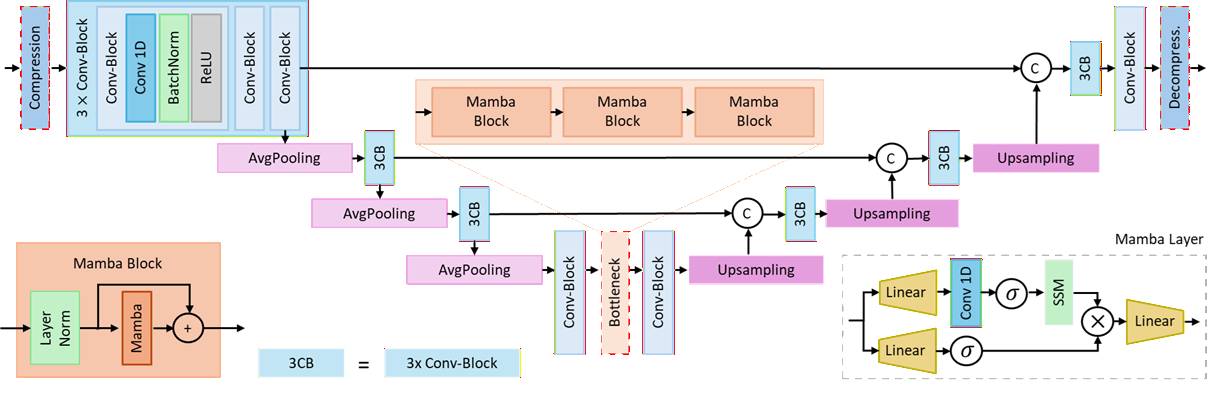}
  \caption{Architecture of UNet-Mamba1-3B. Three Mamba blocks, each preceded by layer normalization, go into the convolutional bottleneck to capture temporal dependencies beyond the ${\sim}1.5$\,s receptive field of the encoder. The Mamba layer follows~\cite{gu2024mamba}.}
  \label{fig:architecture}
\end{figure*}
\subsubsection{Selective State-Space Bottleneck}

To extend the effective receptive field to the full sequence length, we insert three Mamba blocks~\cite{gu2024mamba} at the UNet bottleneck. At this stage, the signal has been down-sampled by $20\times$, so the blocks operate on a 1D sequence of length $\lfloor T/20 \rfloor$ with 48 channels. Each block applies, in sequence: (i) layer normalization over the 48-dimensional feature vector at each time step; (ii) Mamba-1 layer~\cite{gu2024mamba}, which (a) projects the input from dimension $d{=}48$ to $d_E{=}192$ via two parallel linear maps (main and gate branches) with expansion factor $E{=}4$, (b) applies a 1D depthwise convolution of width $d_{\text{conv}}{=}4$ on the main branch for local context, (c) runs a selective state-space model with state dimension $d_{\text{state}}{=}256$, whose transition parameters $(\mathbf{B}, \mathbf{C}, \Delta)$ are computed from each input token, enabling time-varying propagation or forgetting of information along the sequence, (d) element-wise multiplies the SSM output with the SiLU-activated gate branch, and (e) projects back to 48 dimensions; and (iii) residual addition of the block input to the output.
The optional MLP sub-layer ($d_{\text{intermediate}}{=}0$) is omitted, as preliminary experiments showed no improvement in reconstruction while adding parameters. Similar experiments also showed that 3 Mamba blocks yield the best results; we denote the concrete models as \emph{UNet-Mamba1-3B}. See Figure~\ref{fig:architecture} for an overview over the architecture.

\subsubsection{Two-Stage Detail Restoration}
Following~\cite{Qiu2021ECGDenoising}, we optionally cascade the UNet-Mamba with a detail restoration network (DR-net). The DR-net receives a two-channel input---the concatenation of the noisy ECG $\tilde{\mathbf{x}}$ and the first-stage estimate $\hat{\mathbf{x}}$---and produces a refined single-channel output. It comprises four parallel branches (kernel sizes 3, 5, 13, 15) with residual blocks and channel attention, operating at full resolution to preserve sharp features that may be attenuated by the bottleneck. We denote the full two-stage model as \emph{DR-net-Mamba}, and the specific variant with 3 Mamba-Blocks (found to yield optimal performance in preliminary experiments) as \emph{DR-net-Mamba1-3B}.

\subsubsection{Learnable Signal Compression} \label{sec:compression}
We apply a learnable log-compression to the network input: $
    f_\alpha(x) = \operatorname{sign}(x) \cdot \log(1 + \alpha\,|x|),$
where $\alpha > 0$ is a trainable scalar. The inverse transform $f_\alpha^{-1}$ is applied to the network output to recover the original amplitude scale. By factoring out the dynamic range, the compression layer allows the model to devote capacity to morphological structure rather than amplitude variation. Further, compressing the input range may stabilize gradient flow through the recurrent Mamba layers since learning long-range dependencies is prone to exploding and vanishing gradients even with input-dependent recurrence matrices~\cite{bengio1994learning, pascanu2013difficultytrainingrecurrentneural}. See Supplementary Material~\ref{app:compression} (Figure~\ref{fig:compression}) for ablations exploring this benefit.

\subsubsection{Parameter Count and Memory Scaling} \label{sec:params}
The three Mamba blocks add a total of 533{,}088 parameters to the 227K-parameter UNet backbone, yielding a total of 759{,}826 parameters, which is still smaller than our biggest baseline (MECGE, with 819K parameters).
Critically, Mamba is applied only along the time axis after $20\times$ down-sampling, keeping the effective batch size constant regardless of input length. By contrast, MECGE's frequency-axis Mamba branch creates one sequence per time frame, causing effective batch size to scale linearly with signal length and making long-recording training prohibitively expensive (see Supplementary Material~\ref{app:memory}, Figure~\ref{fig:memory}).

\subsection{Training}

All models are trained with MSE loss (batch size 32), normalized per lead using the median and interquartile range, with early stopping on validation loss. Details on training for baseline models are in Supplementary Material~\ref{app:training_baselines} (Table~\ref{tab:hyperparameters}); convergence is confirmed in Supplementary Material~\ref{app:convergence} (Figure~\ref{fig:training_combined}).

\label{sec:lead_training}
For the 12-lead PTB-XL dataset, we deploy UNet-Mamba in two modes: a single \emph{lead-agnostic} model trained across all leads, and 12 separate \emph{lead-specific} 
(or \emph{LS}, for short) 
instances, one per lead. The lead-agnostic variant enables direct comparison with prior work, which is overwhelmingly lead-agnostic by design~\cite{Qiu2021ECGDenoising, antczak2019deeprecurrentneuralnetworks, Chiang_DAE}. 
The single lead is selected by retaining the channel with the fewest detected peaks, as leads with fewer peaks are less likely to be corrupted by noise~\cite{dias}. 
The lead-specific variant allows each instance to specialize to the distinct morphological characteristics of its respective lead---e.g. the dominant R-wave often in V$_5$ versus the predominantly negative QRS complex in aVR~\cite{goldberger2024clinical, ramirez2024art}.

\section{Cohort}


\subsection{Clinical Cohorts}

\textbf{PTB-XL}~\cite{wagner2020ptb} is a large publicly available dataset of 21,799 clinical 12-lead ECGs from 18,869 patients, each 10\,s long at a native sampling frequency of 500\,Hz. Records flagged with baseline drift, static noise, burst noise, or electrode problems are excluded following~\cite{dias}, yielding
13{,}480 (from the original 17{,}418) training and 1{,}639 (from 2{,}183) test segments.

Each record carries a subset of 44 diagnostic statements provided by up to two cardiologists, which can be aggregated into five \emph{superdiagnostic} classes: NORM (normal), MI (myocardial infarction), STTC (ST/T change), CD (conduction disturbance), and HYP (hypertrophy). We leverage these annotations for downstream evaluation of diagnostic classification. Folds~1--8 are used for training, fold~9 for validation, and fold~10 for testing. Further cohort demographics and acquisition details are provided in~\cite{wagner2020ptb} and Supplementary Material~\ref{app:ptbxl_details}.

\textbf{European ST-T Database}~\cite{taddei1992european} consists of 90 annotated excerpts of two-hour ambulatory ECG recordings from 79 subjects in whom myocardial ischemia was diagnosed or suspected. Each record contains two channels sampled at 250\,Hz. The 90 records are split into 54 training, 18 test, and 18 evaluation records.
Within each record, segments are extracted at evenly spaced positions using non-overlapping windows: 2,048 segments for training and 256 segments for testing and evaluation. 

\subsection{Synthetic Cohort}

We generate single-channel synthetic ECGs segments using the dynamical ODE model of~\cite{mcsharry2003dynamical}. Heart rate is drawn uniformly from $[60, 80]$\,bpm for each segment, and the PQRST morphology parameters (amplitudes $a_i$ and widths $b_i$) are sampled uniformly around reference values (Supplementary Material~\ref{app:synth_params}) to introduce inter-segment variability. Each segment is simulated for 45\,s at 360\,Hz; the first 5\,s are discarded to eliminate transient warm-up artifacts, yielding 40\,s effective signals. We generate 1,024 training, 256 test and 256 validation segments. Deterministic random seeds per split ensure reproducibility and non-overlapping populations.

\subsection{Data Extraction and Preprocessing}

All clinical recordings are resampled to 360\,Hz via polyphase resampling to match the frequency expected by the denoising models, then bandpass-filtered (1--45\,Hz), retaining the clinically relevant ECG frequency content---the bulk of QRS spectral energy lies below 40\,Hz~\cite{thakor1984qrs}---while attenuating baseline wander below 1\,Hz and suppressing powerline interference and high-frequency noise above 45\,Hz. Signals are then robustly normalized using the median and interquartile range computed on the training set.

\subsection{Noise Model} \label{sec:noise_model}

\paragraph{Noise source.}
Realistic noise signals are sourced from the MIT-BIH Noise Stress Test Database (NSTDB)~\cite{moody1984noise}, which provides three half-hour, two-channel recordings of noise typical in ambulatory ECG settings: baseline wander~(BW), muscle/EMG artifact~(MA), and electrode motion artifact~(EM). The first channel of each record is split as: 0--15\,mins for training, 15--22.5 for testing, and 22.5--30 for evaluation.

\paragraph{Online noise application.}
During training, all four noise types (BW, MA, EM, and additive white Gaussian noise) are applied simultaneously to each clean ECG segment. For each structured noise type, a random excerpt of the required length is extracted from the corresponding noise bank and scaled to a prescribed SNR. AWGN is generated independently and scaled likewise. Following the ranges recommended by~\cite{HU2024105504}, we set SNR levels to 2.5\,dB for BW, 7.5\,dB for MA, 12.5\,dB for EM, and 22.5\,dB for AWGN. The simultaneous application of all four noise sources ensures that the denoiser is trained under realistic multi-source contamination rather than idealized single-noise conditions. The noise pipeline is validated by comparing empirical SNR distributions against theoretical expectations (Supplementary Material~\ref{app:noise_validation}, Table~\ref{tab:noise_configs}, Figure~\ref{fig:noise_validation}).

\section{Results}

\subsection{Experiments and Evaluation}

\paragraph{Evaluation Metrics.} We quantify reconstruction fidelity using SNR as
$
\mathrm{SNR}(\hat{\mathbf{x}}, \mathbf{x}) =
10 \log_{10}\!\left(
\frac{\left\lVert \mathbf{x} \right\rVert_2^2}
{\left\lVert \mathbf{x} - \hat{\mathbf{x}} \right\rVert_2^2}
\right),$ and RMSE as
$\mathrm{RMSE}(\hat{\mathbf{x}}, \mathbf{x}) =
\sqrt{\frac{1}{T}\left\lVert \hat{\mathbf{x}} - \mathbf{x} \right\rVert_2^2},$
where $\mathbf{x}$ is the clean signal, $\hat{\mathbf{x}}$ the denoised estimate, and $T$ the number of time steps. Each metric is computed per sequence and averaged across all test sequences. For downstream classification we report macro-averaged AUROC (ranking-based), BCE and Brier score (calibration-sensitive, after temperature scaling~\cite{guo2017calibration}), and macro sensitivity, specificity, and F1 
(threshold 0.5). 
For each record and class, we compute a per-entry score (the BCE term, or the squared error for Brier) on temperature-scaled probabilities, then take the flat mean over all entries. These are per-element averages, not summed or prevalence-weighted (micro-averaging).

All metrics use 95\% bootstrap confidence intervals (1,000 resamples, percentile method) following~\cite{strodthoff2021deeplearningecganalysis}.

\paragraph{Downstream Evaluation Pipeline.} \label{sec:downstream_pipeline}
For \textbf{reconstruction (SNR/RMSE)}, one lead per record is selected (fewest detected peaks \cite{dias}) and the denoiser is trained and evaluated on these single-lead signals. The lead-agnostic model pools all leads and is lead-blind, while the lead-specific variant trains 12 separate lead-aware instances. 
To assess \textbf{clinical utility}, clean 12-lead PTB-XL ECGs are separated into individual leads, independently normalized and corrupted with noise, denoised by each model under test, de-normalized, and reassembled into 12-lead signals. Two independent classifiers---Inception1D \cite{Ismail_Fawaz_2020} and ResNet1D-Wang \cite{wang2016timeseriesclassificationscratch} (Supplementary Material~\ref{app:classifiers})---trained exclusively on clean data then classify the reconstructed signals across 44 diagnostic classes. 
We reuse the architectures, hyper-parameters, and training scripts from the PTB-XL benchmark of \cite{strodthoff2021deeplearningecganalysis} (\url{https://github.com/helme/ecg_ptbxl_benchmarking}) without modification, including the standard fold split and multi-label BCE-with-logits objective. We add temperature scaling (\cite{guo2017calibration}): a single global scalar $T$ is fit by minimizing BCE on the clean-condition logits and labels of the validation fold and reused for noisy and denoised conditions. 
This pipeline is illustrated in Figure~\ref{fig:ds_eval}.

\paragraph{Inference Cost.} On a single 10\,s ECG window at 360\,Hz (batch=1, NVIDIA RTX A5000, PyTorch 2.9 + CUDA 12.8, mean of 50 forwards after 10 warmups), DR-net Mamba1-3B runs in 11.9\,ms with 15.6\,MB peak GPU memory, faster than DR-net-IMUNet (16.0\,ms) at comparable memory (Table~V). All evaluated models process 10\,s of signal in under 50\,ms, comfortably real-time for clinical deployment.

\subsection{ECG Reconstruction}

\begin{figure*}[t]
    \centering
    \includegraphics[width=0.9\linewidth]{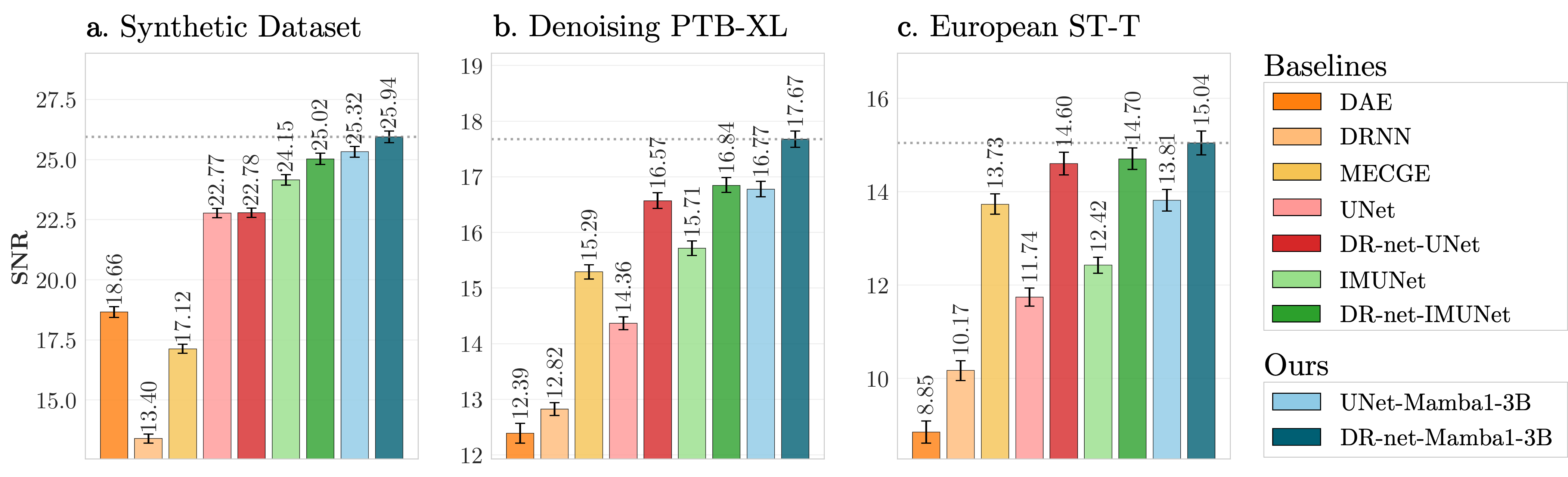}
    \caption{Reconstruction SNR (dB) on synthetic (a), PTB-XL (b), and European ST-T (c) datasets. Empirical noisy SNR = 0.96\,dB (medium noise setting, see Supplementary Material~\ref{app:noise_validation}). Analogous RMSE results are reported in Figure~\ref{fig:performance_rmse} in Supplementary Material~\ref{app:performance_rmse}.}
    \label{fig:performance}
\end{figure*}

Figures~\ref{fig:performance} and~\ref{fig:performance_rmse} report SNR and RMSE, respectively, across all three datasets under the medium noise setting (combined input SNR $\approx$ 0.96\,dB; see Supplementary Material~\ref{app:noise_validation}). DR-net-Mamba achieves the highest SNR and lowest RMSE on all three datasets. On the synthetic benchmark (Figure~\ref{fig:performance}a), UNet-Mamba and DR-net-Mamba obtain approximately 2.5--3\,dB higher SNR than their base models (UNet and DR-net-UNet), while DRNN, DAE, and MECGE consistently yield the weakest performance.

Performance rankings on PTB-XL (Figure~\ref{fig:performance}b) and European ST-T (Figure~\ref{fig:performance}c) are largely consistent with the synthetic benchmark. On clinical data, the performance gap between Stage~1 and Stage~2 models is more pronounced. The relationship between training data volume and reconstruction performance is explored in Supplementary Material~\ref{app:data_volume}.

To confirm the reported improvements are not artifacts of overlapping per-model CIs, we ran paired bootstrap tests of the mean difference $\Delta = \bar{x}_A - \bar{x}_B$ between every baseline and our models for each of the three datasets (for MECGE, the 1800-sample windows were stitched back into the 14400-sample recordings before computing the metric). For each pair, recording indices were resampled with replacement $B=1000$ times, using the same indices for both models per iteration to preserve pairing, and 95\% percentile confidence intervals of $\Delta$ are reported. A CI excluding 0 indicates a significant difference at $\alpha = 0.05$.

We find that every comparison of our models against 
their respective baseline yields a significant difference (at $\alpha = 0.05$) on each of the three datasets: For the synthetic test set ($n=256$ source recordings), the smallest gain is $+0.29$ dB SNR over DR-net-IMUNet with a CI of $[+0.15, +0.44]$. On PTB-XL (
${n=1{,}639}$ recordings, 10\,s each at 360\,Hz), DR-net Mamba1-3B outperforms every baseline (smallest gain $+0.83$ dB SNR over DR-net-IMUNet, CI $[+0.78, +0.88]$), while on the European ST-T database ($n=256$ recordings, 40\,s) the smallest gain of $+0.34$ dB SNR, CI $[+0.20, +0.49]$ is achieved over DR-net-IMUNet. 
The individual results are given in Tables~\ref{tab:pairwise_synth}, \ref{tab:pairwise_ptbxl} and \ref{tab:pairwise_stt} in Section~\ref{app:performance_rmse_pairwise} in the Supplementary Material.

The compression layer is a potential confounder. We have therefore run ablation studies to separate the effects of Mamba, DR-net, and log compression. We isolated the three components by evaluating a 6-cell grid on European ST-T (UNet baseline, with/without DR-net, with/without Mamba, with/without learnable compression) on the same $N=256$ test recordings. All pairwise SNR differences are computed with a paired bootstrap: recording indices are resampled with replacement $B=1000$ times using the same indices for both models per iteration, yielding the 95\% CI of $\Delta = \bar{x}_A - \bar{x}_B$. Mamba and DR-net each contribute large, significant gains; learnable compression adds +0.9 dB SNR on top of single-stage Mamba (CI [+0.7, +1.0]) but its effect vanishes once DR-net is added (-0.0 dB, CI [-0.1, +0.1]), indicating that compression and the two-stage DR-net design capture overlapping benefits.
The detailed numbers are tabulated in Table~\ref{tab:compression_impact} in the Supplementary Material.

\subsection{Noise Robustness}

\begin{figure*}[t]
    \centering
    \includegraphics[width=0.9\linewidth]{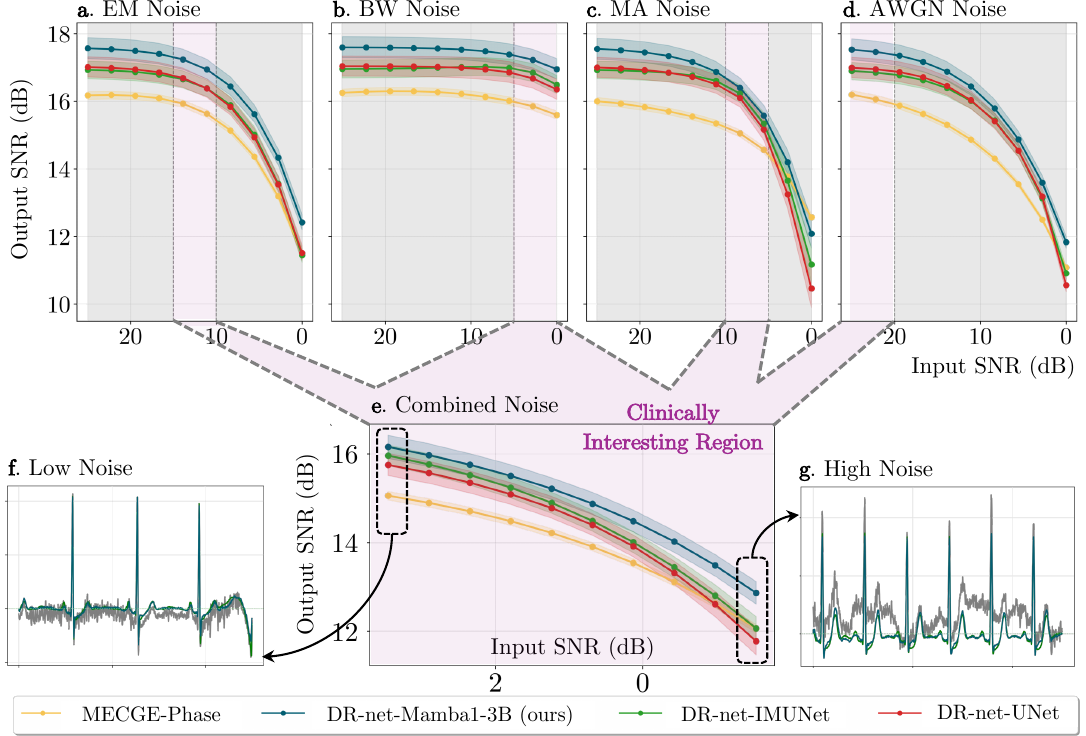}
    \caption{Reconstruction SNR (dB) as a function of input SNR for each noise type (EM, BW, MA, AWGN) and the combined condition on European ST-T. Grey-shaded areas denote input SNR values outside clinically plausible ranges~\cite{HU2024105504}. Shaded bands indicate 95\% bootstrap CIs.}
    \label{fig:noise_robustness}
\end{figure*}

Figure~\ref{fig:noise_robustness} reports reconstruction SNR as a function of input SNR for each noise type individually and for the combined condition on the European ST-T dataset, restricted to the best-performing models for clarity. Grey-shaded regions indicate input SNR values outside the plausible clinical operating ranges as reported by~\cite{HU2024105504}. DR-net-Mamba achieves the highest output SNR across all noise types and clinically relevant input SNR levels. The advantage over purely convolutional models is most pronounced under heavy AWGN. MECGE shows notable robustness under heavy muscle artifact and AWGN. Under the combined noise condition, both DR-net-Mamba and MECGE degrade more gracefully than DR-net-UNet and DR-net-IMUNet.

\subsection{Impact of Recording Length}

\begin{figure}[t]
    \centering
    \includegraphics[width=\linewidth]{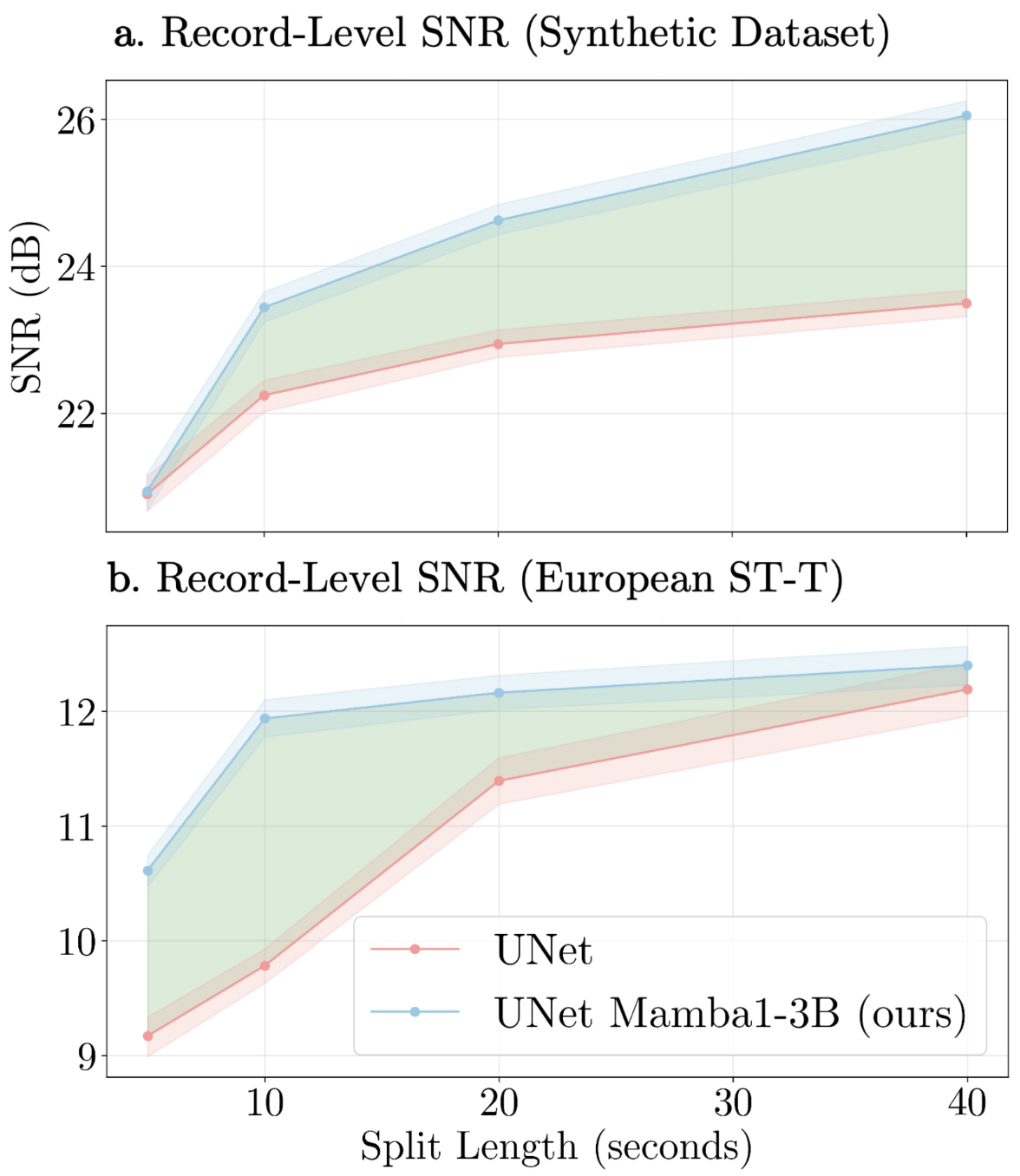}
    \caption{Reconstruction SNR (dB) as a function of recording length for UNet and UNet-Mamba (without signal compression), on synthetic (a) and European ST-T (b). Shaded bands indicate 95\% bootstrap CIs.}
    \label{fig:length}
\end{figure}

Figure~\ref{fig:length} compares UNet and UNet-Mamba (without log compression, to isolate the bottleneck contribution) as a function of recording length using a curriculum-based protocol. 
Models are trained in four stages of increasing sequence length (5\,s $\to$ 10\,s $\to$ 20\,s $\to$ 40\,s), with the original 40\,s long records being cut into shorter windows of 7{,}200, 3{,}600, and 1{,}800 samples (20\,s, 10\,s, and 5\,s), respectively, alongside the full 40\,s. Each model is thus trained on the same total recording length, but a varying maximal length. In the training, each model is initialized using the weights from the previous stage, 
with compression disabled so the two architectures differ only by the presence of the Mamba bottleneck. On the synthetic dataset (Figure~\ref{fig:length}a), the two models achieve comparable SNR at 5\,s, but UNet-Mamba consistently outperforms UNet for longer recordings, with the gain growing from approximately 1\,dB at 10\,s to roughly 2.5\,dB at 40\,s. On the European ST-T dataset (Figure~\ref{fig:length}b), UNet-Mamba already outperforms UNet by approximately 1\,dB at 5\,s. The improvement peaks around 10\,s and then diminishes for longer sequences, in contrast to the monotonic growth observed on synthetic data. Training details for each curriculum stage are reported in Supplementary Material~\ref{app:length_training} (Tables~\ref{tab:training_length_syn}--\ref{tab:training_length_eu}).

\subsection{Clinical Translation: Downstream Disease Prediction}

\subsubsection{Aggregated Evaluation} \label{sec:downstream_aggregated}

\begin{figure*}[thbp]
    \centering
    \includegraphics[width=1\linewidth]{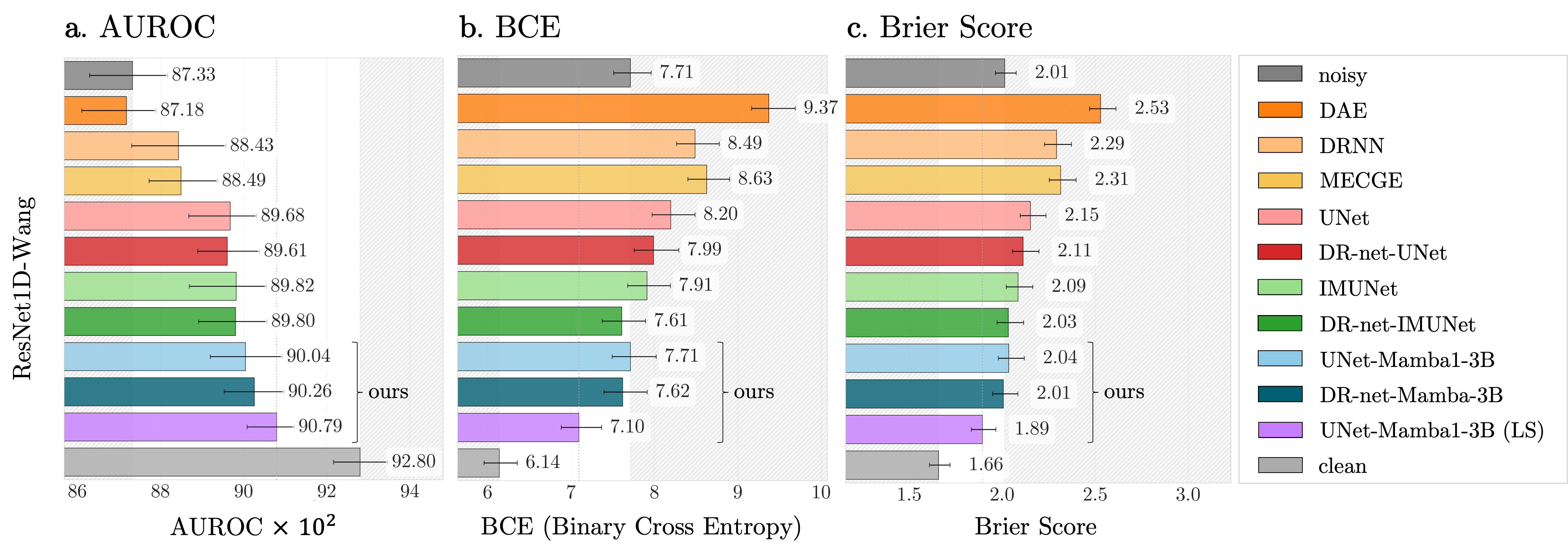}
    \caption{Downstream diagnostic classification (macro and superdiagnostic AUROC) on PTB-XL using ResNet1D-Wang, comparing denoised signals against clean and noisy baselines (strong noise setting).}
    \label{fig:downstream_high}
\end{figure*}

Figure~\ref{fig:downstream_high} summarizes downstream classification performance on PTB-XL under the strong noise setting (BW 0$dB$; MA 5$dB$; EM 10$dB$; AWGN 20$dB$) using the ResNet1D-Wang backbone--see Figure \ref{fig:downstream_high_inception1D} in Supplementary Material \ref{app:macro_inception} for the Inception1D backbone. Results under a lower noise setting are reported in Supplementary Material~\ref{app:downstream_medium}. Noise reduces AUROC by $0.069$ for Inception1D and $0.055$ for ResNet1D-Wang. The Mamba-based models achieve the best overall AUROC across both classifiers and consistently improve over their base models. The lead-specific UNet-Mamba (Section~\ref{sec:lead_training}) further improves AUROC on ResNet1D-Wang. 

The calibration picture is more nuanced and backbone-dependent. Among the denoisers, the Mamba-based models attain the best BCE and Brier scores on both classifiers, but whether denoising improves calibration \emph{over the noisy input} depends on the classifier. On Inception1D, denoising broadly helps: every model that improves AUROC also lowers BCE and Brier relative to the noisy baseline, with the lead-specific UNet-Mamba best on both (BCE $6.89$ and Brier $1.82$ vs.\ noisy $7.85$/$2.02$). On ResNet1D-Wang the effect is weaker: several denoisers---including some Mamba variants---sit at or slightly above the noisy baseline on Brier even when they improve BCE, and only the lead-specific UNet-Mamba clearly improves \emph{both} BCE and Brier over the noisy input (BCE $7.10$ and Brier $1.89$ vs.\ noisy $7.71$/$2.01$). Thus the lead-specific variant is the only denoiser to improve calibration over the noisy baseline on both classifiers, while other models trade improved discrimination against degraded or unchanged calibration. We return to this discrimination--calibration gap in Section~\ref{sec:downstream_medium_disc}.

To understand the clinical relevance of these results, Table~\ref{tab:combined_perclass_macro_inception} reports macro-averaged metrics for Inception1D (strong noise, bold/underlined: best/second best). The noisy signal achieves higher sensitivity than the clean baseline ($0.852$ vs.\ $0.815$) at the cost of substantially lower specificity ($0.697$ vs.\ $0.863$). The three Mamba variants rank first through third on F1, combining specificity of $0.839$--$0.841$ with sensitivity of $0.789$--$0.808$. Similar trends hold for ResNet1D-Wang (Supplementary Material~\ref{app:macro_wang}, Table~\ref{tab:macro_f1_wang}).

\subsubsection{Evaluation by Superdiagnostic Class}

Table~\ref{tab:combined_perclass_macro_inception} additionally breaks down AUROC by superdiagnostic class. The Mamba-based models achieve the highest AUROC across all five classes. The lead-specific UNet-Mamba leads in HYP, MI, NORM, and STTC, while DR-net-Mamba scores second-best in CD, HYP, MI, and STTC. Results under a lower noise setting (Tables~\ref{tab:per_class_default}--\ref{tab:macro_f1_inception_medium}) and per-class sensitivity, specificity, and F1 (Tables~\ref{tab:per_class_f1_inception}--\ref{tab:per_class_f1_wang}) are reported in Appendices~\ref{app:downstream_medium} and~\ref{app:superdiagnostic_spec}.

\subsubsection{Evaluation by Diagnostic Class}

Figure~\ref{fig:tree_inception}a isolates the effect of adding a Mamba bottleneck to the UNet while keeping all other design choices fixed. UNet-Mamba consistently outperforms UNet on STTC (ST/T change) and its subclasses, whereas UNet achieves comparable or slightly better AUROC on CD (conduction disturbance) and most of its subclasses.

Figure~\ref{fig:tree_inception}b shows the absolute AUROC of the lead-specific UNet-Mamba per diagnostic class, with color encoding improvement over the noisy baseline. For all diagnostic classes with more than 18 segments, denoising either improves or preserves classification performance. The corresponding breakdowns for ResNet1D-Wang are reported in Supplementary Material~\ref{app:downstream_wang} (Figure~\ref{fig:tree_wang}) and show similar trends.


A natural alternative to the denoise-plus-clean-classifier pipeline is to train the classifier directly on noise-corrupted ECGs. As shown in Table~\ref{tab:auc_cleanNoisy} (strong noise, $SNR \approx -1.5\,dB$) , this noise-aware classifier provides a strong baseline, achieving an AUROC of 0.923 compared with 0.909 for the denoise-plus-clean-classifier pipeline, which therefore does not outperform direct noise-aware training in downstream discrimination. This comparison, however, favors the noise-aware classifier in an important respect: it is trained and evaluated using the same noise family and noise setting, such that the noisy test data remain in-distribution. Moreover, this strategy requires the classifier to be retrained for each noise condition and each downstream classification task. In contrast, our approach decouples signal restoration from downstream prediction: the denoiser is trained once for the target noise setting and can subsequently be reused with different off-the-shelf classifiers trained on clean ECGs and for different downstream tasks. Importantly, the denoiser also reconstructs the underlying ECG signal itself, rather than learning only a noise-robust decision boundary for a specific set of labels.

\begin{table}[htbp]
  \centering
  \caption{Downstream classifier AUROC under strong noise.
}
  \label{tab:auc_cleanNoisy}
  \begin{tabular}{llc}
    \toprule
    Classifier trained on & Test input                         & AUC \\
    \midrule
    Clean                 & Clean                              & $0.925\ [0.916, 0.932]$ \\
    Clean                 & Noisy                              & $0.879\ [0.865, 0.892]$ \\
    Noisy                 & Noisy                              & $0.923\ [0.913, 0.930]$ \\ 
    Clean                 & \makecell[l]{Denoised\\(Mamba1-3B Lead Aware)} & $0.909\ [0.901, 0.918]$ \\
    \bottomrule
  \end{tabular}
\end{table}


\section{Discussion}
The central finding of this work is that augmenting a convolutional encoder--decoder with a selective state-space bottleneck consistently improves reconstruction fidelity and noise robustness, and improves downstream diagnostic \emph{discrimination} (AUROC), while its effect on \emph{calibration} (BCE, Brier) is smaller and classifier-dependent---and that the nature of these effects is informative about when and why long-range temporal context matters for physiological signal processing.

\subsection{Long-Range Context as an Inductive Bias for Physiological Denoising}

\begin{table*}[t]
\centering
\caption{Per-superdiagnostic-class AUROC and overall macro-averaged sensitivity, specificity, and F1 on PTB-XL 
}
\label{tab:combined_perclass_macro_inception}
\small
\resizebox{\textwidth}{!}{%
\begin{tabular}{lcccccccc}
\toprule
 & CD & HYP & MI & NORM & STTC & \multicolumn{3}{c}{Overall} \\
\cmidrule(lr){2-6} \cmidrule(lr){7-9}
 & \multicolumn{5}{c}{AUROC} & Sensitivity & Specificity & F1 \\
\midrule
UNet-Mamba1-3B & \textbf{.925 $\pm$ .01} & .883 $\pm$ .02 & .909 $\pm$ .02 & .926 $\pm$ .01 & .880 $\pm$ .02 & .789 $\pm$ .02 & \underline{.840 $\pm$ .01} & .703 $\pm$ .02 \\
DR-net-Mamba1-3B & \underline{.921 $\pm$ .01} & \underline{.887 $\pm$ .02} & \underline{.909 $\pm$ .02} & .926 $\pm$ .01 & \underline{.881 $\pm$ .02} & \underline{.794 $\pm$ .02} & \textbf{.841 $\pm$ .01} & \underline{.706 $\pm$ .02} \\
UNet-Mamba1-3B (LS) & .920 $\pm$ .01 & \textbf{.887 $\pm$ .02} & \textbf{.910 $\pm$ .01} & \textbf{.928 $\pm$ .01} & \textbf{.887 $\pm$ .02} & \textbf{.808 $\pm$ .02} & .839 $\pm$ .01 & \textbf{.715 $\pm$ .02} \\
\midrule
DRNN & .900 $\pm$ .02 & .885 $\pm$ .02 & .895 $\pm$ .02 & .910 $\pm$ .02 & .824 $\pm$ .02 & .783 $\pm$ .02 & .780 $\pm$ .01 & .651 $\pm$ .02 \\
DAE & .860 $\pm$ .02 & .854 $\pm$ .03 & .874 $\pm$ .01 & .907 $\pm$ .01 & .848 $\pm$ .02 & .709 $\pm$ .02 & .819 $\pm$ .01 & .638 $\pm$ .02 \\
UNet & .914 $\pm$ .02 & .864 $\pm$ .03 & .904 $\pm$ .01 & .921 $\pm$ .01 & .865 $\pm$ .02 & .764 $\pm$ .01 & .822 $\pm$ .01 & .672 $\pm$ .02 \\
DR-net-UNet & .914 $\pm$ .01 & .875 $\pm$ .03 & .904 $\pm$ .01 & .921 $\pm$ .01 & .866 $\pm$ .02 & .784 $\pm$ .02 & .828 $\pm$ .01 & .690 $\pm$ .01 \\
IMUNet & .914 $\pm$ .01 & .875 $\pm$ .02 & .901 $\pm$ .01 & .925 $\pm$ .01 & .879 $\pm$ .02 & .767 $\pm$ .02 & \underline{.840 $\pm$ .01} & .691 $\pm$ .02 \\
DR-net-IMUNet & .912 $\pm$ .01 & .883 $\pm$ .03 & .900 $\pm$ .02 & \underline{.927 $\pm$ .01} & .879 $\pm$ .02 & .785 $\pm$ .02 & \textbf{.841 $\pm$ .01} & .699 $\pm$ .02 \\
MECGE & .905 $\pm$ .01 & .870 $\pm$ .03 & .887 $\pm$ .02 & .917 $\pm$ .01 & .872 $\pm$ .02 & .790 $\pm$ .02 & .808 $\pm$ .01 & .680 $\pm$ .02 \\
\midrule
clean & .932 $\pm$ .01 & .906 $\pm$ .02 & .925 $\pm$ .02 & .938 $\pm$ .01 & .914 $\pm$ .02 & .815 $\pm$ .02 & .863 $\pm$ .01 & .738 $\pm$ .01 \\
noisy & .903 $\pm$ .02 & .872 $\pm$ .03 & .875 $\pm$ .02 & .911 $\pm$ .02 & .863 $\pm$ .02 & .852 $\pm$ .02 & .697 $\pm$ .01 & .632 $\pm$ .02 \\
\bottomrule
\end{tabular}%
}
\end{table*}

The recording-length experiment (Figure~\ref{fig:length}) provides a clear direct evidence of the above. On synthetic data, where every beat shares the same parametric PQRST morphology~\cite{mcsharry2003dynamical}, the Mamba advantage grows monotonically with sequence length because the state of distant beats is genuinely predictive of the current one. The UNet's convolutional receptive field saturates at ${\sim}1.5$\,s, so additional temporal context cannot be exploited by the bottleneck convolutions, whereas the Mamba blocks propagate information across the full downsampled sequence. On clinical data, the Mamba advantage peaks around 10\,s and then diminishes, reflecting the lower temporal regularity of real ECGs: inter-beat variability, heart rate variability, and pathological waveform heterogeneity make distant beats less predictive than in the stationary synthetic case. This contrast offers a practical guideline: selective state-space bottlenecks are most beneficial when the signal exhibits structured temporal regularity beyond the convolutional receptive field, a condition that holds for many physiological recordings (e.g., respiratory signals, EEG rhythms, continuous glucose monitoring) but whose degree is application-specific.

\begin{figure*}[t]
    \centering
    \includegraphics[width=\linewidth]{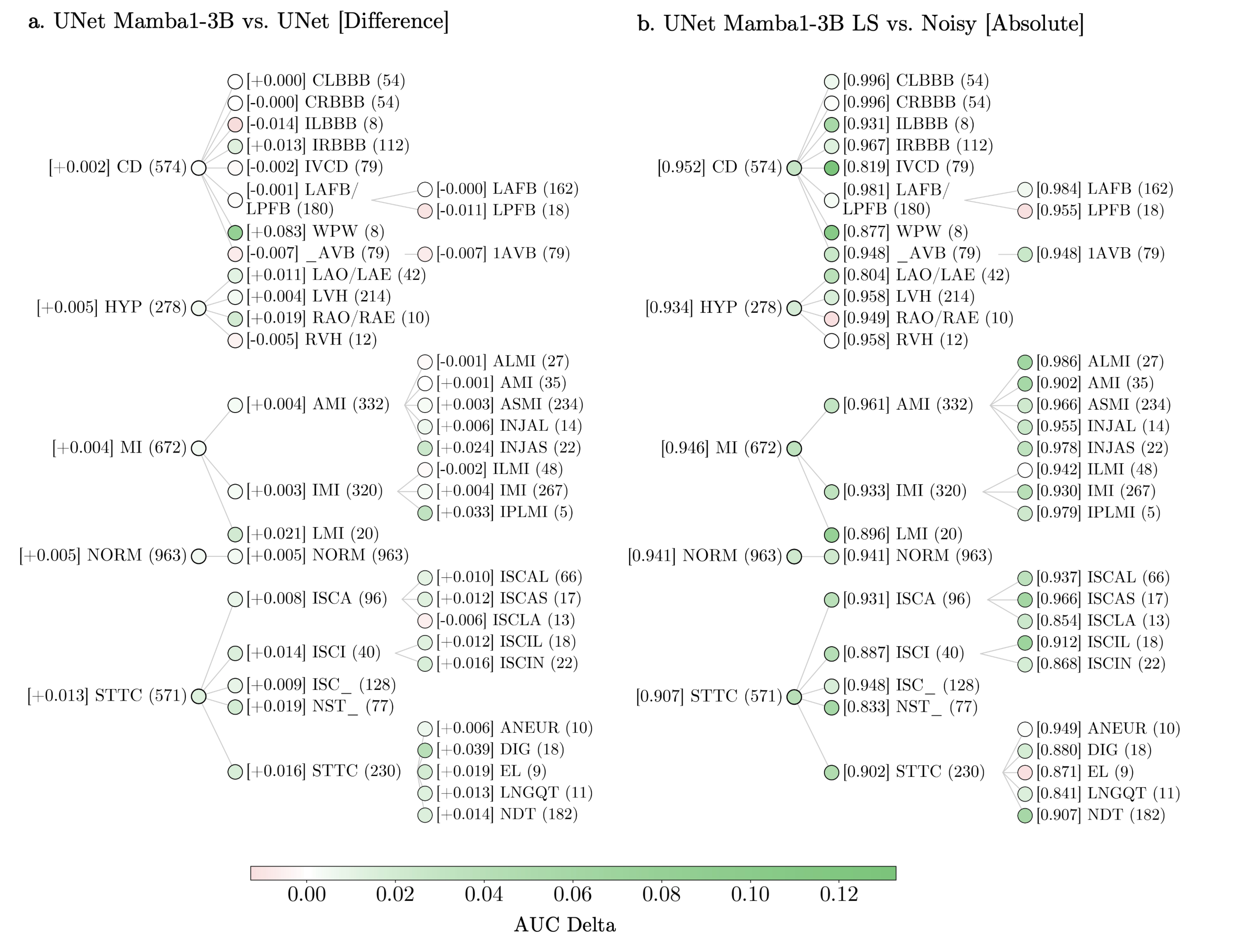}
    \caption{\textbf{Per-diagnostic-class downstream AUROC (Inception1D).} \textbf{(a)} AUROC delta when adding a Mamba bottleneck to the UNet, all other design choices held constant. \textbf{(b)} Absolute AUROC of the lead-specific UNet-Mamba; color indicates improvement over the noisy baseline. In both panels, diagnostic classes with fewer than five segments are omitted. See Supplementary Material~\ref{app:abbreviations} for abbreviations.}
    \label{fig:tree_inception}
\end{figure*}

The noise robustness results (Figure~\ref{fig:noise_robustness}) reinforce this interpretation. The Mamba advantage over purely convolutional models is most pronounced under heavy AWGN. Structured noise sources (BW, MA, EM) carry recognizable temporal patterns that convolutional models can learn to suppress within their receptive field~\cite{kumar2020detection}. AWGN is sample-wise independent and offers no such local cues; as its intensity grows, reconstruction must rely on temporal predictability across beats---precisely the form of context the Mamba bottleneck provides. Under the combined noise condition, both DR-net-Mamba and MECGE degrade more gracefully than purely convolutional models, suggesting that global context---whether temporal (Mamba) or spectral (MECGE; \cite{hung2024mecgemambabasedecgenhancer})---is beneficial when multiple noise sources interact simultaneously.

\subsection{Clinical Translation: When Long-Range Context Helps Most}\label{sec:downstream_medium_disc}

The downstream evaluation confirms that Mamba-based denoising improves diagnostic performance across both classifiers and nearly all pathology classes. Crucially, the per-class analysis (Figure~\ref{fig:tree_inception}a, Table~\ref{tab:sttc_vs_cd}) reveals that this benefit is morphology-dependent, offering actionable insight into which clinical scenarios gain most from long-range temporal modeling. The Mamba bottleneck consistently and substantially improves AUROC on STTC diagnoses---slow, broad waveforms (200--400\,ms) whose interpretation depends on the surrounding baseline~\cite{surawicz2008chou} and that are easily confounded with structured noise such as baseline wander. 
For conduction disturbances, diagnosed from QRS-internal intervals such as the R-wave peak time (45--60\,ms)~\cite{surawicz2009aha}, performance matches the convolutional baseline: with $20\times$ downsampling these span barely one bottleneck time step, leaving little room for improvement. Table~\ref{tab:sttc_vs_cd} summarizes this resolution-versus-context trade-off. Importantly, for all diagnostic classes with more than 18 segments, the lead-specific Mamba model either improves or preserves AUROC relative to the noisy baseline (Figure~\ref{fig:tree_inception}b), confirming that denoising does not harm downstream classification at the individual-diagnosis level.

The calibration results further complicate the reconstruction-to-utility mapping, and they differ sharply between the two classifiers. On Inception1D, denoising generally improves calibration: models that raise AUROC also lower BCE and Brier relative to the noisy baseline. On ResNet1D-Wang, however, this coupling breaks down---several denoisers improve AUROC yet sit at or above the noisy baseline on Brier, and some worsen both BCE and Brier. A plausible explanation is that on ResNet1D-Wang denoising helps the classifier on borderline cases (improving rankings and thus AUROC) but reduces confidence on non-borderline segments, where the denoised signal may differ subtly from what the classifier was trained on; because BCE and Brier directly penalize reduced confidence on every segment, this loss can outweigh the gains on the borderline subset. Across both classifiers, the lead-specific Mamba variant is the only denoiser that improves both BCE and Brier over the noisy baseline, likely because per-lead specialization yields denoised signals closer to the clean morphology on which the classifiers were trained. This contrast highlights the importance of evaluating denoising not only through discrimination metrics (AUROC) but also through calibration metrics (BCE, Brier)---and of doing so per classifier---when the downstream application involves clinical decision-making with probabilistic outputs.

\subsection{The Sensitivity--Specificity Shift Induced by Noise}

The observation that noisy signals yield \emph{higher} sensitivity than clean baselines ($0.852$ vs.\ $0.815$ for Inception1D) while substantially degrading specificity ($0.697$ vs.\ $0.863$) reveals a clinically important operating-point shift. Noise-corrupted morphology makes many negative examples resemble positive ones, pushing the classifier toward a higher false-positive rate. Denoising partially reverses this shift: all models reduce sensitivity relative to the noisy input while recovering specificity. Mamba models achieve the best balance, with the highest F1 scores and specificity closest to the clean baseline. For clinical deployment, the specificity recovery offered by denoising may be more valuable than the raw AUROC improvement, particularly in screening settings where false positives trigger costly follow-up.

\subsection{DR-net-Mamba in the Broader Denoising Landscape}

Prior ECG denoising studies~\cite{Qiu2021ECGDenoising, Chiang_DAE, antczak2019deeprecurrentneuralnetworks} each use different clean-signal sources, noise protocols, segment lengths, and metric definitions, making cross-study comparison of absolute values unreliable. Our noise pipeline differs fundamentally: each source is scaled to per-type SNR targets drawn from clinically calibrated ranges~\cite{HU2024105504}, and all four types are applied simultaneously, yielding evaluation conditions more challenging than those in the baseline studies. Rather than comparing absolute metric values across incompatible setups, we focus on within-study rankings using a unified noise protocol applied consistently to all models. The absence of a standardized ECG denoising benchmark remains an open problem for the field.

\subsection*{Further Machine Learning in Healthcare Insights}

This work offers three insights that extend beyond ECG denoising. First, we demonstrate that inserting selective state-space blocks at the bottleneck of a convolutional encoder--decoder is a parameter-efficient strategy for injecting long-range temporal context into physiological signal models, with the benefit growing with recording length. Second, we show that reconstruction gains do not automatically transfer to downstream clinical utility: the relationship between denoising quality and diagnostic accuracy is morphology-dependent, with context-sensitive waveforms (e.g., ST/T changes) benefiting most from long-range modeling while sharp features (e.g., QRS complexes) can be harmed by bottleneck over-smoothing. Third, we find that lead-specific denoising---training separate models per ECG lead---can preserve calibration where lead-agnostic models degrade it, highlighting that evaluation beyond discrimination is essential for clinical deployment.

\subsection{Limitations} Various limitations arise in this work: (i) \emph{Ground-truth quality.} Neither PTB-XL nor European ST-T provides truly clean references. Residual artifacts survive preprocessing and become part of the reconstruction target through the MSE loss. Reconstruction metrics therefore partially measure reproduction of these artifacts rather than genuine signal recovery, and Stage~2 improvements on clinical data should be interpreted with this caveat. (ii) \emph{Single-lead pipeline.} Each lead is denoised independently and reassembled for downstream classification. Inter-lead spatial correlations in the 12-lead ECG are not exploited. A multi-lead architecture could improve both reconstruction and diagnostic performance, and would avoid the $12\times$ training cost of the lead-specific variant. (iii) \emph{Noise model scope.} All models are trained and evaluated with four noise types at prescribed SNR levels. Other clinically relevant artifact types (e.g., lead misplacement, pacemaker spikes, motion artifacts in wearable devices) and adaptive or non-stationary noise levels are not covered. (iv) \emph{Downstream evaluation scope.} The downstream pipeline uses classifiers trained on clean data and evaluated on denoised data. In practice, classifiers may be trained on noisy data or jointly optimized with the denoiser. Our pipeline isolates the effect of denoising but does not capture potential benefits from end-to-end training. (v) \emph{Recording-length scaling on clinical data.} The Mamba advantage peaks around 10\,s on clinical recordings and then diminishes, unlike the monotonic growth on synthetic data. Whether this reflects intrinsic limits of long-range context in irregular clinical signals or artifacts of the training protocol (e.g., boundary effects~\cite{pielawski2020hann}) warrants further investigation.


\section*{Conflicts of Interest}

S. R.-C. reports equity, consulting, and intellectual property with Physcade Inc. Disclosures are unrelated to this work. All other authors have nothing to disclose.



\renewcommand*{\bibfont}{\small}
\printbibliography

\newpage

\appendix
\setcounter{figure}{0}
\setcounter{table}{0}
\renewcommand{\thefigure}{S\arabic{figure}}
\renewcommand{\thetable}{S\arabic{table}}
\renewcommand{\theHfigure}{supp.\arabic{figure}}
\renewcommand{\theHtable}{supp.\arabic{table}}


\etocdepthtag.toc{appendix}

\begingroup
\etocsettagdepth{main}{none}
\etocsettagdepth{appendix}{subsection}
\etocsettocstyle{\noindent\textbf{Supplementary Material Overview}\medskip\par\noindent\hrulefill\medskip}{\medskip\noindent\hrulefill\bigskip}
{\small\tableofcontents}
\endgroup

\section{Supplementary Methods}

\subsection{Receptive Field Computation} \label{app:receptive}

The receptive field of a chain of $L$ convolutions is the number of input samples that can influence a single output sample. For a network where layer $l$ has kernel size $k_l$ and stride $s_l$:
\begin{equation}
    r_0 = \sum_{l=1}^{L} \left( (k_l - 1) \prod_{i=1}^{l-1} s_i \right) + 1.
\end{equation}
Pooling layers are treated identically, using their pool size as $k_l$ and pool stride as $s_l$. Table~\ref{tab:receptive_field} details the per-layer contribution for the UNet encoder.

\begin{table}[htbp]
\centering
\caption{Per-layer receptive field computation for the UNet encoder.}
\label{tab:receptive_field}
\small
\begin{tabular}{@{}clccrc@{}}
\toprule
$l$ & Layer Type & $s_l$ & $k_l$ & $\displaystyle\prod_{i=1}^{l-1} s_i$ & $\displaystyle(k_l - 1)\prod_{i=1}^{l-1} s_i$ \\
\midrule
1  & Conv     & 1 & 25 &  1 & 24 \\
2  & Conv     & 1 & 25 &  1 & 24 \\
3  & Conv     & 1 & 25 &  1 & 24 \\
4  & AvgPool  & 5 &  5 &  1 &  4 \\
5  & Conv     & 1 & 15 &  5 & 70 \\
6  & Conv     & 1 & 15 &  5 & 70 \\
7  & Conv     & 1 & 15 &  5 & 70 \\
8  & AvgPool  & 2 &  2 &  5 &  5 \\
9  & Conv     & 1 &  5 & 10 & 40 \\
10 & Conv     & 1 &  5 & 10 & 40 \\
11 & Conv     & 1 &  5 & 10 & 40 \\
12 & AvgPool  & 2 &  2 & 10 & 10 \\
13 & Conv     & 1 &  3 & 20 & 40 \\
14 & Conv     & 1 &  3 & 20 & 40 \\
15 & Conv     & 1 &  3 & 20 & 40 \\
\midrule
\multicolumn{5}{r}{Sum:} & 541 \\
\multicolumn{5}{r}{\textbf{Receptive Field} ($r_0 = \text{Sum} + 1$):} & \textbf{542} \\
\bottomrule
\end{tabular}
\end{table}

\subsection{Signal Compression Ablation} \label{app:compression}

\begin{figure}[t]
  \centering
  \includegraphics[width=0.8\linewidth]{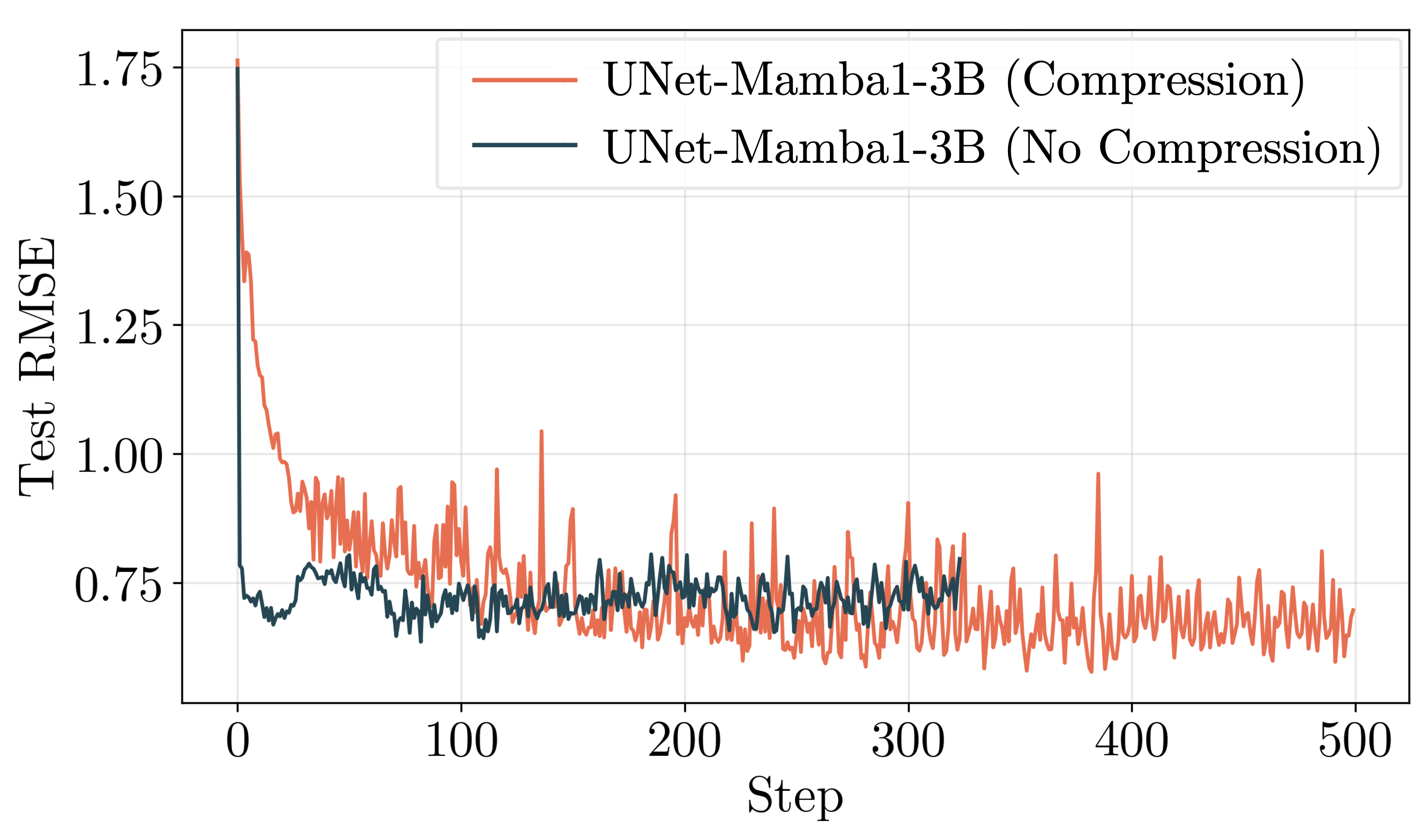}
  \caption{Test RMSE over training steps for UNet-Mamba with and without the learnable log-compression layer. Removing compression yields faster initial convergence but a worse final optimum.}
  \label{fig:compression}
\end{figure}

Figure~\ref{fig:compression} compares the training dynamics of UNet-Mamba with and without the learnable log-compression layer (Section~\ref{sec:compression}). Without compression, the model converges faster initially---consistent with easily fitting the dominant scale component---but plateaus at a higher error, indicating that capacity is spent on amplitude scale at the expense of morphological structure.

\subsection{Memory Scaling Comparison} \label{app:memory}

\begin{figure}[t]
  \centering
  \includegraphics[width=\linewidth]{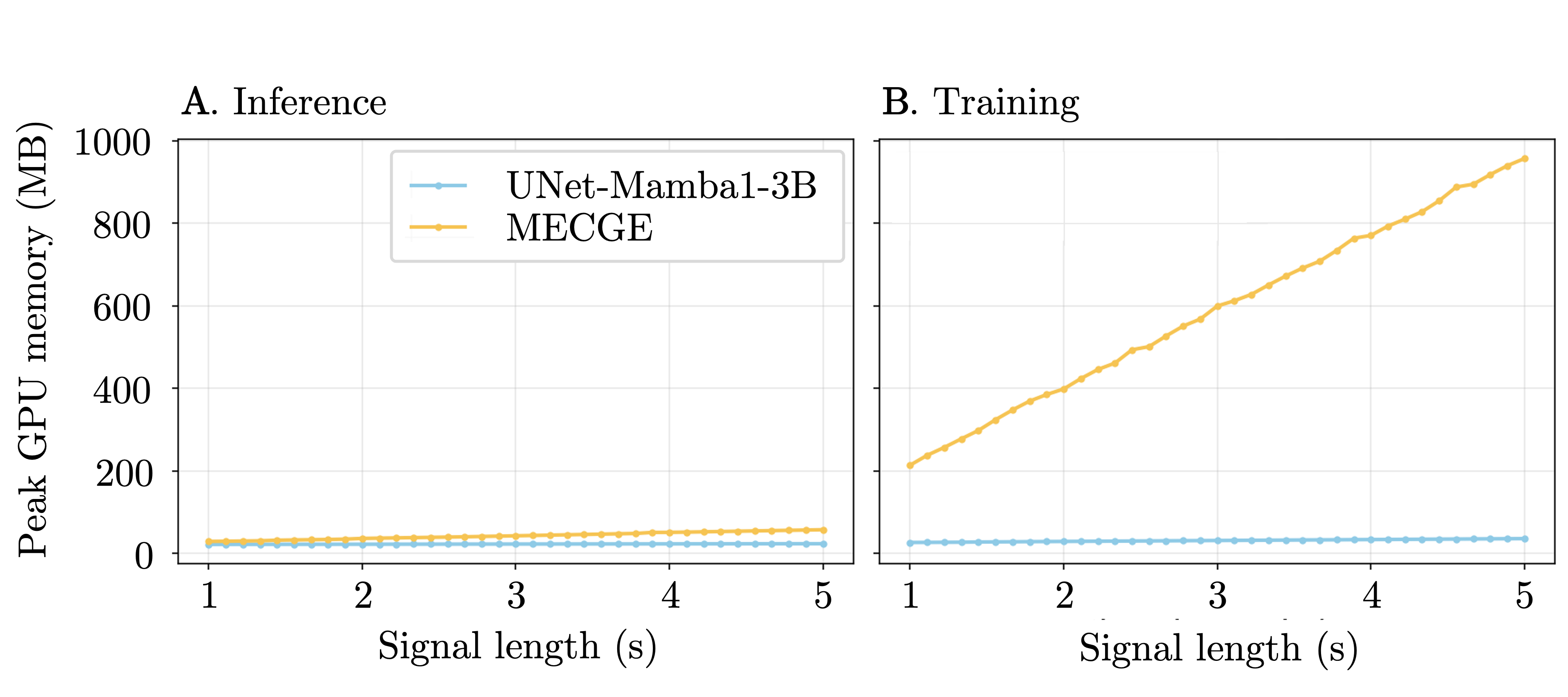}
  \caption{Peak GPU memory allocation as a function of ECG signal length (1--5\,s at 360\,Hz, batch size 2). \textbf{A}: inference (forward pass). \textbf{B}: training (forward + backward). MECGE training requires substantially more memory than UNet-Mamba due to STFT-domain processing.}
  \label{fig:memory}
\end{figure}

MECGE~\cite{hung2024mecgemambabasedecgenhancer} applies bidirectional Mamba along both time and frequency axes in the STFT domain. Its Frequency Bi-Mamba branch creates one sequence per time frame, so the effective batch size scales as $\mathcal{O}(B \cdot T / h)$ where $h$ is the STFT hop size. For a 40\,s recording at 360\,Hz with hop size $h{=}8$, this yields an effective batch of $B \times 1{,}800$ in the frequency branch alone. By contrast, UNet-Mamba applies Mamba only along the time axis after $20\times$ down-sampling, keeping the effective batch size at $B$ and the sequence length at $T/20$. Figure~\ref{fig:memory} compares peak GPU memory allocation.

\subsection{Training Details for Baseline Models}
\label{app:training_baselines}

For the baseline models (UNet, IMUNet, DAE, DRNN, DR-net variants), the optimizer and scheduler follow the respective original publications, using Adam with learning rate $10^{-3}$ and ReduceLROnPlateau. MECGE uses AdamW with learning rate $10^{-4}$ and exponential decay. For UNet-Mamba, we use Adam with $\eta_{\max} = 8 \times 10^{-4}$ and cosine annealing with 10-epoch linear warm-up from $\eta_{\max} \times 0.01$ to $\eta_{\max}$, followed by cosine decay over 400 cycles to $\eta_{\min} = 10^{-6}$, following~\cite{gu2024mamba}. Full hyperparameters are listed in Supplementary Material~\ref{app:hyperparameters}.

\subsection{Training Hyperparameters} \label{app:hyperparameters}

Table~\ref{tab:hyperparameters} summarizes the training hyperparameters for all models. All models use MSE loss with batch size 32.

\begin{table}[t]
\centering
\caption{Training hyperparameters for baseline and proposed models.}
\label{tab:hyperparameters}
\small

\resizebox{\columnwidth}{!}{%
\begin{tabular}{lccc}
\toprule
 & UNet / IMUNet / DAE / & MECGE & UNet-Mamba \\
 & DRNN / DR-net-S2 & & (ours) \\
\midrule
Optimizer & Adam & AdamW & Adam \\
Learning rate & $10^{-3}$ & $10^{-4}$ & $8\times10^{-4}$ \\
LR scheduler & ReduceLROnPlateau & Exp.\ decay & Cosine + warmup \\
Warmup epochs & --- & --- & 10 \\
$T_{\max}$ (cosine) & --- & --- & 400 \\
$\eta_{\min}$ & --- & --- & $10^{-6}$ \\
Max epochs & 500 & 120 & 500 \\
ES patience & 120--240 & 15 & 120--240 \\
\bottomrule
\end{tabular}%
}

\end{table}

\subsection{Training Convergence} \label{app:convergence}

\begin{figure}[thbp]
  \centering
  \includegraphics[width=\linewidth]{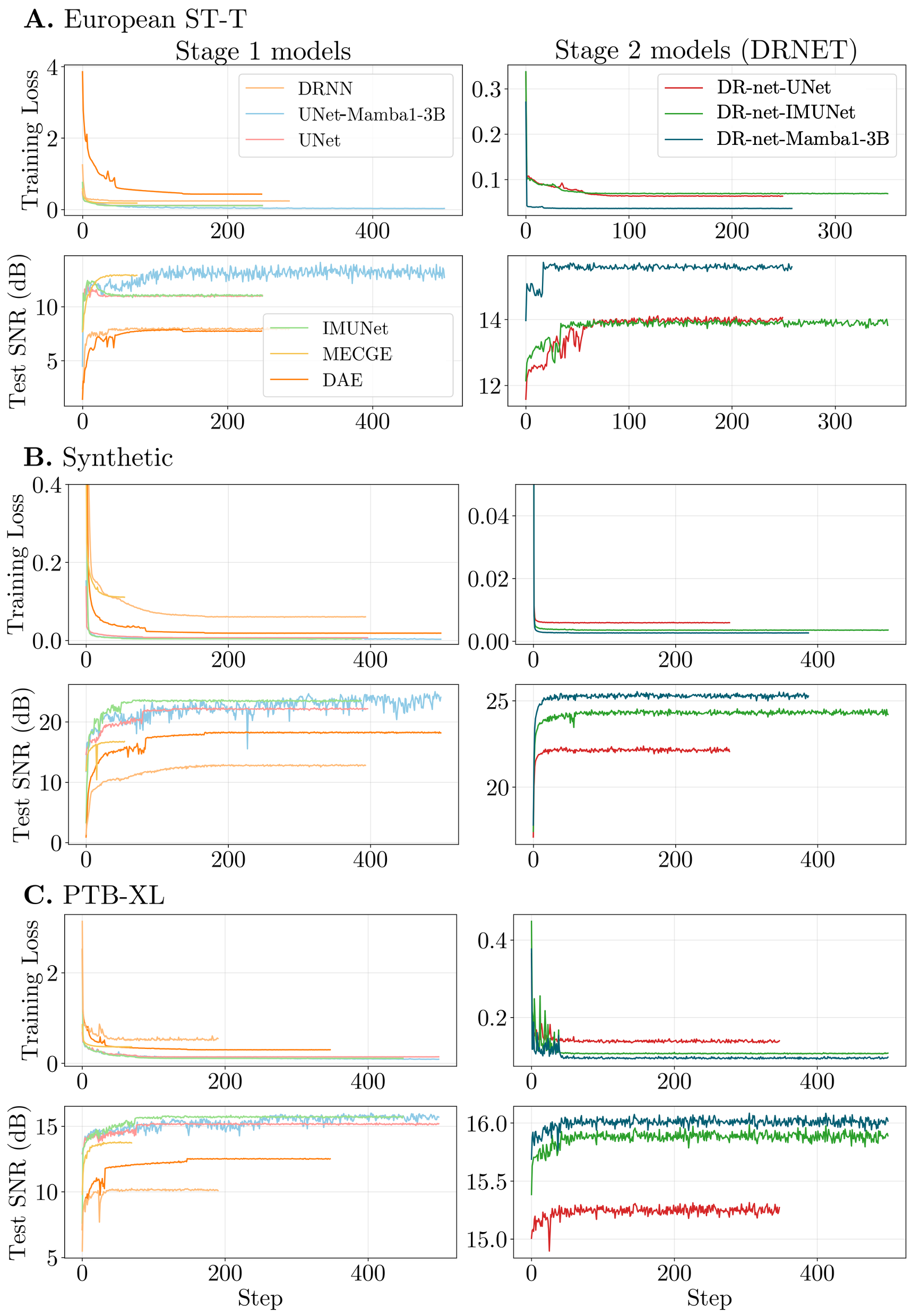}
  \caption{Training loss (top row) and test SNR (bottom row) over training steps across all three datasets: \textbf{(A)} European ST-T, \textbf{(B)} synthetic, \textbf{(C)} PTB-XL. Stage~1 models on the left of each panel; Stage~2 (DR-net) on the right.}
  \label{fig:training_combined}
\end{figure}

All models converge well before the maximum number of epochs, with test SNR curves reaching a stable plateau. Figure~\ref{fig:training_combined} shows the train loss and test SNR across all three datasets.

\section{Supplementary Materials}

\subsection{PTB-XL} \label{app:ptbxl_details}
The waveform data underlying PTB-XL were collected with devices from Schiller AG at the Physikalisch-Technische Bundesanstalt over nearly seven years (October 1989--June 1996). The cohort is 52\% male and 48\% female, with ages spanning 0--95 years (median 62, interquartile range 22). Signals are stored in WFDB format at 16-bit precision with a resolution of 1\,$\mu$V/LSB. Each record was annotated with a free-text report converted into standardized SCP-ECG statements; a large fraction was additionally validated by a second cardiologist. The recommended 10-fold splits are stratified by patient. Records in folds~9 and~10 underwent at least one human evaluation and are therefore of particularly high label quality. For full details, see~\cite{wagner2020ptb}.

\subsection{Synthetic ECG Parameters} \label{app:synth_params}

Table~\ref{tab:synth_params} lists the morphology parameters used for synthetic ECG generation, following the notation of~\cite{mcsharry2003dynamical}. For each waveform event $i \in \{P, Q, R, S, T\}$, the amplitude $a_i$ is sampled uniformly from $[c_{a_i} - \delta_{a_i},\, c_{a_i} + \delta_{a_i}]$ and the width $b_i$ from $[c_{b_i} - \delta_{b_i},\, c_{b_i} + \delta_{b_i}]$.

\begin{table}[htbp]
\centering
\caption{Synthetic ECG simulation parameters for the PQRST morphology.}
\label{tab:synth_params}
\small
\begin{tabular}{lccccc}
\toprule
 & P & Q & R & S & T \\
\midrule
$c_{a_i}$ & 1.2 & $-5.0$ & 30.0 & $-7.5$ & 0.75 \\
$\delta_{a_i}$ & 0.6 & 0.2 & 0.0 & 1.0 & 0.35 \\
$c_{b_i}$ & 0.25 & 0.1 & 0.1 & 0.1 & 0.4 \\
$\delta_{b_i}$ & 0.1 & 0.1 & 0.0 & 0.0 & 0.0 \\
\bottomrule
\end{tabular}
\end{table}

\subsection{Noise Pipeline Validation} \label{app:noise_validation}

We validated the noise injection pipeline by loading 1,000 preprocessed PTB-XL segments and applying the full noise pipeline under three configurations (light, medium, strong), corresponding to the upper bound, midpoint, and lower bound of the per-type SNR ranges recommended by~\cite{HU2024105504}. The theoretical combined SNR when $K$ independent noise sources are added simultaneously is:
\begin{equation}
    \mathrm{SNR}_{\mathrm{combined}} = 10 \log_{10} \left( \frac{1}{\sum_{k=1}^{K} 10^{-\mathrm{SNR}_k / 10}} \right).
\end{equation}

\begin{figure*}[htbp]
  \centering
  \includegraphics[width=\linewidth]{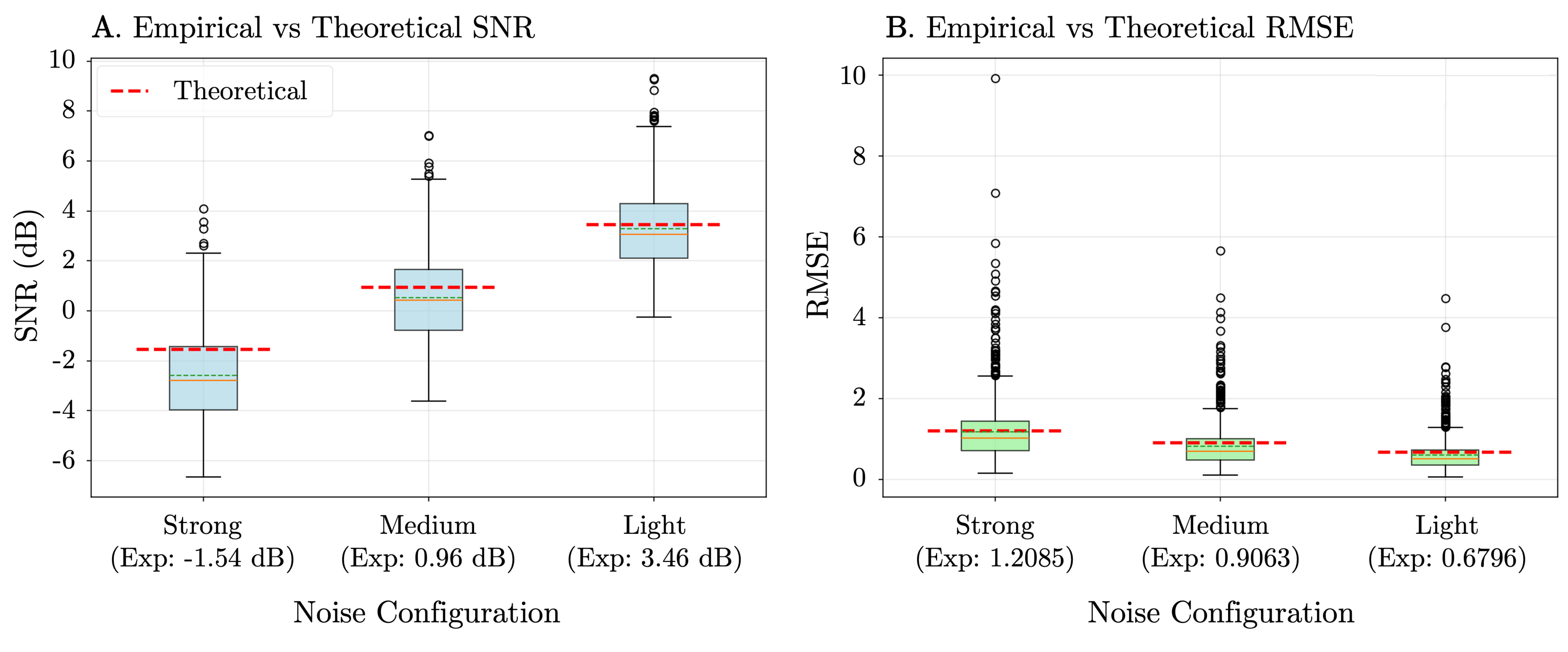}
  \caption{Empirical vs.\ theoretical SNR (left) and RMSE (right) across three noise configurations. Boxplots show per-segment distributions; dashed lines indicate theoretical expected values.}
  \label{fig:noise_validation}
\end{figure*}

Table~\ref{tab:noise_configs} summarizes the per-type SNR values and predicted combined SNR for each configuration. Figure~\ref{fig:noise_validation} shows that the empirical medians closely match the theoretical predictions and the ordering across configurations is preserved.

\begin{table}[htbp]
\centering
\caption{Noise configurations. Per-type SNR values (dB) are drawn from ranges recommended by~\cite{HU2024105504}.}
\label{tab:noise_configs}
\small
\begin{tabular}{lccccc}
\toprule
Config & BW & MA & EM & AWGN & Combined SNR \\
\midrule
Light   & 5.0 & 10.0 & 15.0 & 25.0 & 3.46 \\
Medium & 2.5 &  7.5 & 12.5 & 22.5 & 0.96 \\
Strong  & 0.0 &  5.0 & 10.0 & 20.0 & $-1.54$ \\
\bottomrule
\end{tabular}
\end{table}

\section{Supplementary Results}

\subsection{Downstream Classifier Architectures} \label{app:classifiers}

The two downstream classification models are:

\textbf{Inception1D}: a 1D adaptation of InceptionTime~\cite{Ismail_Fawaz_2020}, stacking six inception blocks with multi-scale parallel convolutions and residual shortcuts every three blocks, followed by adaptive concatenation pooling and a linear head.

\textbf{ResNet1D-Wang}: a 1D residual network following~\cite{wang2016timeseriesclassificationscratch}, with three single-block residual stages using kernel sizes 5 and 3, stride-1 convolutions, and the same adaptive concatenation pooling and linear head as Inception1D.

Both models are trained using binary cross-entropy loss on clean data from the PTB-XL training folds and accept 12-channel inputs with an independent binary prediction per diagnostic class.

\subsection{Recording-Length Experiment: Training Details} \label{app:length_training}

\begin{figure}[htbp]
    \centering
    \includegraphics[width=\linewidth]{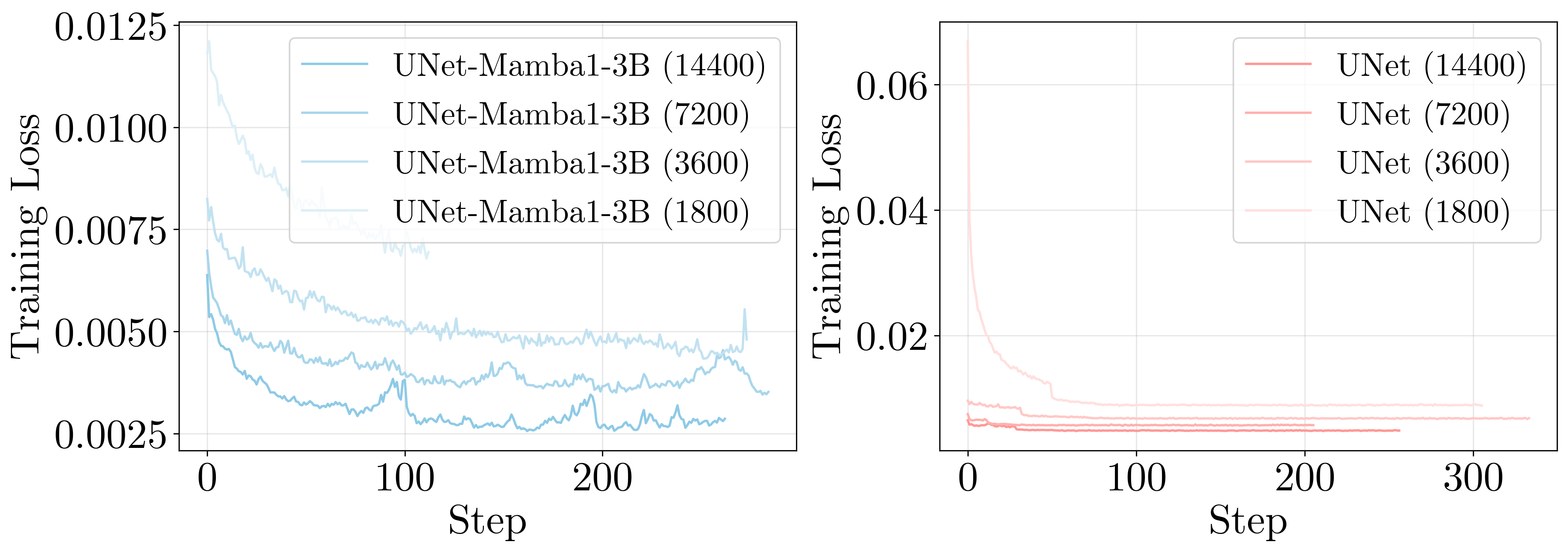}
    \caption{Train loss curves for the recording-length experiment on the synthetic dataset.}
    \label{fig:length_loss_syn}
\end{figure}

\begin{figure}[htbp]
    \centering
    \includegraphics[width=\linewidth]{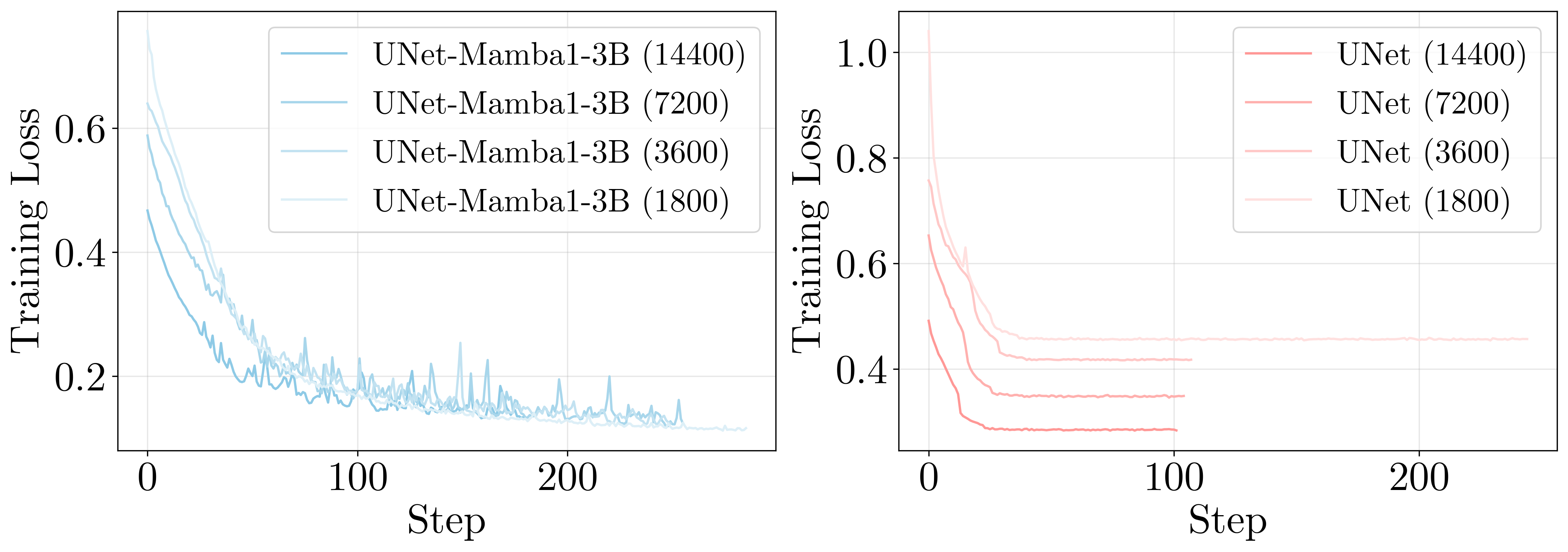}
    \caption{Train loss curves for the recording-length experiment on the European ST-T dataset.}
    \label{fig:length_loss_eu}
\end{figure}

Tables~\ref{tab:training_length_syn} and~\ref{tab:training_length_eu} summarize the number of epochs and wall-clock training time per curriculum stage on the synthetic and European ST-T datasets, respectively. Figures~\ref{fig:length_loss_syn} and~\ref{fig:length_loss_eu} show the corresponding train loss curves.

\begin{table}[hbtp]
  \centering
  \caption{Training details for the synthetic dataset across split lengths. Trained on a single NVIDIA RTX A5000.}
  \begin{tabular}{lccc}
    \toprule
    \textbf{Method} & \textbf{Split Length} & \textbf{Epochs} & \textbf{Runtime} \\
    \midrule
    UNet-Mamba1-3B & 14400 & 180 & 41m 45s \\
    UNet-Mamba1-3B & 7200  & 256 & 1h 4m 52s \\
    UNet-Mamba1-3B & 3600  & 250 & 1h 50m 39s \\
    UNet-Mamba1-3B & 1800  & 286 & 7h 18m 38s \\
    \midrule
    UNet           & 14400 & 102 & 12m 33s \\
    UNet           & 7200  & 105 & 21m 22s \\
    UNet           & 3600  & 108 & 38m 26s \\
    UNet           & 1800  & 245 & 5h 42m 15s \\
    \bottomrule
  \end{tabular}
  \label{tab:training_length_syn}
\end{table}

\begin{table}[t]
  \centering
  \caption{Training details for the European ST-T dataset across split lengths. Trained on a single NVIDIA RTX A5000.}
  \begin{tabular}{lccc}
    \toprule
    \textbf{Method} & \textbf{Split Length} & \textbf{Epochs} & \textbf{Runtime} \\
    \midrule
    UNet-Mamba1-3B & 14400 & 263 & 32m 13s \\
    UNet-Mamba1-3B & 7200  & 285 & 39m 44s \\
    UNet-Mamba1-3B & 3600  & 274 & 1h 7m 7s \\
    UNet-Mamba1-3B & 1800  & 113 & 1h 34m 27s \\
    \midrule
    UNet           & 14400 & 257 & 17m 13s \\
    UNet           & 7200  & 206 & 21m 56s \\
    UNet           & 3600  & 334 & 1h 4m \\
    UNet           & 1800  & 306 & 3h 32m 46s \\
    \bottomrule
  \end{tabular}
  \label{tab:training_length_eu}
\end{table}

\subsection{Effect of Training Data Volume} \label{app:data_volume}

\begin{figure*}[htbp]
    \centering
    \includegraphics[width=0.7\linewidth]{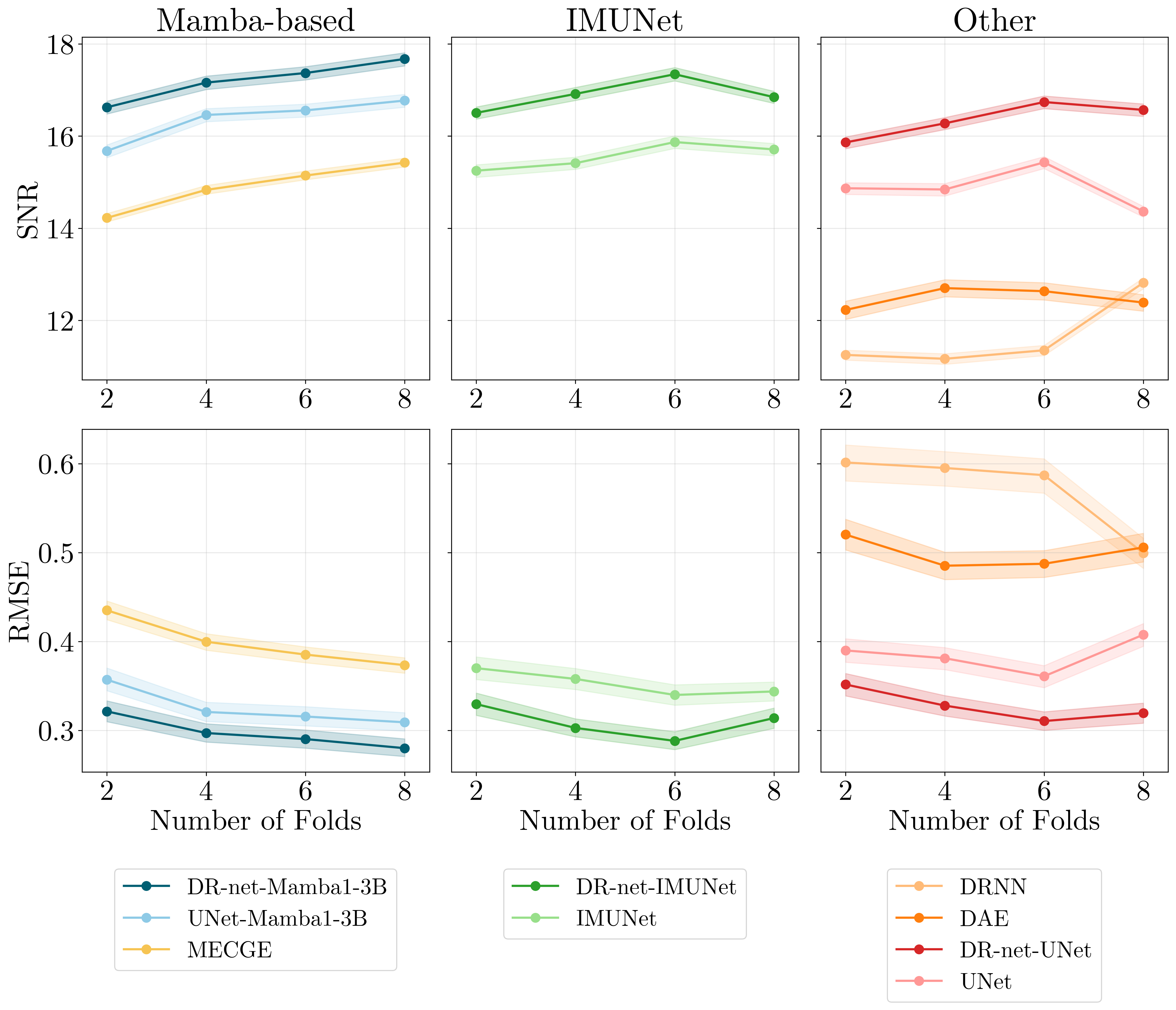}
    \caption{Reconstruction SNR (top) and RMSE (bottom) on PTB-XL as a function of the number of training folds.}
    \label{fig:snr_rmse_vs_folds}
\end{figure*}

Figure~\ref{fig:snr_rmse_vs_folds} shows reconstruction SNR and RMSE on PTB-XL as a function of the number of training folds. Mamba-based models improve monotonically up to 8 folds, whereas UNet and IMUNet peak at 6 folds and slightly degrade with additional data, suggesting that the higher capacity of the Mamba bottleneck allows it to continue benefiting from additional training data where purely convolutional models saturate earlier.

\subsection{Reconstruction RMSE} \label{app:performance_rmse}

\begin{figure*}[thbp]
    \centering
    \includegraphics[width=\linewidth]{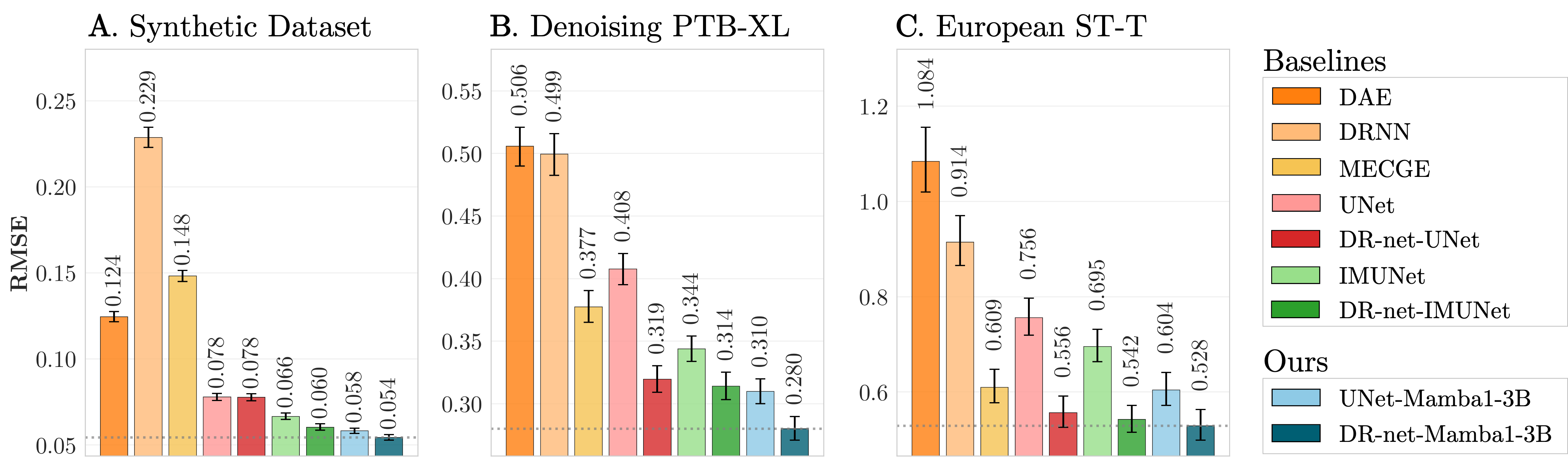}
    \caption{Reconstruction RMSE on synthetic (A), PTB-XL (B), and European ST-T (C) datasets. Empirical noisy SNR = 0.96\,dB (medium noise setting, see Supplementary Material~\ref{app:noise_validation}). This figure supplements Figure~\ref{fig:performance}.}
    \label{fig:performance_rmse}
\end{figure*}

Figure~\ref{fig:performance_rmse} reports reconstruction RMSE across all three datasets, supplementing the SNR results in Figure~\ref{fig:performance}.

\subsection{Pairwise Tests for Reconstruction SNR} \label{app:performance_rmse_pairwise}

Table~\ref{tab:pairwise_synth} shows the pairwise comparison between the proposed models and each baseline model on the synthetic test set along with the 95\% confidence intervals. Table~\ref{tab:pairwise_ptbxl} gives the results of the same evaluation on the PTB-XL dataset, and Table~\ref{tab:pairwise_stt} on the ST-T dataset. On all three datasets, the confidence intervals do not contain 0, implying a significant difference (at $\alpha=0.05$ level) between the baseline models and the proposed models.

\begin{table}[htbp]
  \centering
  \caption{Pairwise comparison with 95\% confidence intervals for the synthetic
  test set ($n=256$ source recordings). DR-net Mamba1-3B outperforms every
  baseline (smallest gain 
  $+0.92$\,dB SNR over DR-net-IMUNet, CI $[+0.79,+1.06]$). 
  See Figure~\ref{fig:performance}a.}
  \label{tab:pairwise_synth}

  \small
  \resizebox{\columnwidth}{!}{%
    \begin{tabular}{lcc}
      \toprule
      Baseline
      & \makecell{UNet\\Mamba1-3B}
      & \makecell{DR-net\\Mamba1-3B} \\
      \midrule
      DRNN
      & $+11.922\ [+11.761,+12.098]$
      & $+12.546\ [+12.355,+12.720]$ \\
      DAE
      & $+6.657\ [+6.492,+6.811]$
      & $+7.281\ [+7.116,+7.440]$ \\
      UNet
      & $+2.551\ [+2.414,+2.699]$
      & $+3.175\ [+3.025,+3.332]$ \\
      DR-net-UNet
      & $+2.538\ [+2.386,+2.682]$
      & $+3.162\ [+3.015,+3.316]$ \\
      IMUNet
      & $+1.172\ [+1.019,+1.309]$
      & $+1.796\ [+1.646,+1.954]$ \\
      DR-net-IMUNet
      & $+0.294\ [+0.148,+0.440]$
      & $+0.918\ [+0.785,+1.060]$ \\
      MECGE
      & $+8.194\ [+8.002,+8.400]$
      & $+8.818\ [+8.623,+9.049]$ \\
      UNet Mamba1-3B
      & ---
      & $+0.624\ [+0.573,+0.674]$ \\
      DR-net Mamba1-3B
      & $-0.624\ [-0.674,-0.573]$
      & --- \\
      \bottomrule
    \end{tabular}%
  }

\end{table}

\begin{table}[htbp]
  \centering
  \caption{Pairwise comparison with 95\% confidence intervals for PTB-XL
  ($n=1{,}639$ recordings, 10\,s each at 360\,Hz). DR-net Mamba1-3B
  outperforms every baseline (smallest gain $+0.83$\,dB SNR over
  DR-net-IMUNet, CI $[+0.78,+0.88]$). See Figure~\ref{fig:performance}b.}
  \label{tab:pairwise_ptbxl}

  \small
  \resizebox{\columnwidth}{!}{%
    \begin{tabular}{lcc}
      \toprule
      Baseline
      & UNet Mamba1-3B
      & DR-net Mamba1-3B \\
      \midrule
      DRNN
      & $+3.955\ [+3.889,+4.019]$
      & $+4.854\ [+4.785,+4.926]$ \\
      DAE
      & $+4.385\ [+4.297,+4.473]$
      & $+5.284\ [+5.193,+5.382]$ \\
      UNet
      & $+2.409\ [+2.359,+2.460]$
      & $+3.308\ [+3.244,+3.366]$ \\
      DR-net-UNet
      & $+0.206\ [+0.162,+0.252]$
      & $+1.105\ [+1.062,+1.148]$ \\
      IMUNet
      & $+1.063\ [+1.020,+1.104]$
      & $+1.962\ [+1.911,+2.016]$ \\
      DR-net-IMUNet
      & $-0.070\ [-0.116,-0.022]$
      & $+0.828\ [+0.778,+0.882]$ \\
      MECGE
      & $+1.485\ [+1.395,+1.579]$
      & $+2.384\ [+2.299,+2.475]$ \\
      UNet Mamba1-3B
      & ---
      & $+0.899\ [+0.868,+0.930]$ \\
      DR-net Mamba1-3B
      & $-0.899\ [-0.930,-0.868]$
      & --- \\
      \bottomrule
    \end{tabular}%
  }
\end{table}

\begin{table}[htbp]
  \centering
  \caption{Pairwise comparison with 95\% confidence intervals for the ST-T
  database ($n=256$ recordings, 40\,s). DR-net Mamba1-3B outperforms every
  baseline (smallest gain $+0.34$\,dB SNR over DR-net-IMUNet,
  CI $[+0.20,+0.49]$). See Figure~\ref{fig:performance}c.}
  \label{tab:pairwise_stt}

  \small
  \resizebox{\columnwidth}{!}{%
    \begin{tabular}{lcc}
      \toprule
      Baseline
      & \makecell{UNet\\Mamba1-3B}
      & \makecell{DR-net\\Mamba1-3B} \\
      \midrule
      DRNN
      & $+3.650\ [+3.489,+3.813]$
      & $+4.878\ [+4.712,+5.055]$ \\
      DAE
      & $+4.966\ [+4.801,+5.113]$
      & $+6.194\ [+6.014,+6.380]$ \\
      UNet
      & $+2.078\ [+1.940,+2.226]$
      & $+3.307\ [+3.157,+3.453]$ \\
      DR-net-UNet
      & $-0.784\ [-0.930,-0.622]$
      & $+0.445\ [+0.315,+0.592]$ \\
      IMUNet
      & $+1.391\ [+1.232,+1.550]$
      & $+2.619\ [+2.428,+2.815]$ \\
      DR-net-IMUNet
      & $-0.887\ [-1.045,-0.713]$
      & $+0.342\ [+0.198,+0.493]$ \\
      MECGE
      & $+0.085\ [-0.085,+0.261]$
      & $+1.313\ [+1.138,+1.472]$ \\
      UNet Mamba1-3B
      & ---
      & $+1.229\ [+1.130,+1.331]$ \\
      DR-net Mamba1-3B
      & $-1.229\ [-1.331,-1.130]$
      & --- \\
      \bottomrule
    \end{tabular}%
  }
\end{table}
\subsection{Impact of Compression Layer in Reconstruction Accuracy}\label{app:compression_impact_rmse}
  
\begin{table}[htbp]
  \centering
  \caption{Ablation study separating the effects of Mamba, DR-net, and log
  compression. Mean differences and 95\% confidence intervals are reported
  from paired bootstrap analysis with $B=1000$ resamples with replacement.}
  \label{tab:compression_impact}

  \scriptsize
  \setlength{\tabcolsep}{2pt}

  \resizebox{\columnwidth}{!}{%
    \begin{tabular}{lcccccc}
      \toprule
      Baseline
        & UNet
        & \makecell{DR-net-\\UNet}
        & \makecell{UNet\\Mamba1-3B\\(No Comp)}
        & \makecell{DR-net\\Mamba1-3B\\(No Comp)}
        & \makecell{UNet\\Mamba1-3B}
        & \makecell{DR-net\\Mamba1-3B} \\
      \midrule

      UNet
        & ---
        & \makecell{$+2.9$\\{\small $[+2.7,+3.0]$}}
        & \makecell{$+1.2$\\{\small $[+1.1,+1.3]$}}
        & \makecell{$+3.3$\\{\small $[+3.2,+3.4]$}}
        & \makecell{$+2.1$\\{\small $[+1.9,+2.2]$}}
        & \makecell{$+3.3$\\{\small $[+3.2,+3.5]$}} \\
      \addlinespace

      \makecell[l]{DR-net-\\UNet}
        & {}
        & ---
        & \makecell{$-1.6$\\{\small $[-1.8,-1.5]$}}
        & \makecell{$+0.5$\\{\small $[+0.3,+0.6]$}}
        & \makecell{$-0.8$\\{\small $[-0.9,-0.6]$}}
        & \makecell{$+0.4$\\{\small $[+0.3,+0.6]$}} \\
      \addlinespace

      \makecell[l]{UNet\\Mamba1-3B\\(No Comp)}
        & {}
        & {}
        & ---
        & \makecell{$+2.1$\\{\small $[+2.0,+2.2]$}}
        & \makecell{$+0.9$\\{\small $[+0.7,+1.0]$}}
        & \makecell{$+2.1$\\{\small $[+1.9,+2.2]$}} \\
      \addlinespace

      \makecell[l]{DR-net\\Mamba1-3B\\(No Comp)}
        & {}
        & {}
        & {}
        & ---
        & \makecell{$-1.2$\\{\small $[-1.4,-1.1]$}}
        & \makecell{$-0.0$\\{\small $[-0.1,+0.1]$}} \\
      \addlinespace

      \makecell[l]{UNet\\Mamba1-3B}
        & {}
        & {}
        & {}
        & {}
        & ---
        & \makecell{$+1.2$\\{\small $[+1.1,+1.3]$}} \\
      \bottomrule
    \end{tabular}%
  }
\end{table}

\subsection{Downstream Classification under Lower Noise} \label{app:downstream_medium}

\begin{figure}[htbp]
    \centering
    \includegraphics[width=\linewidth]{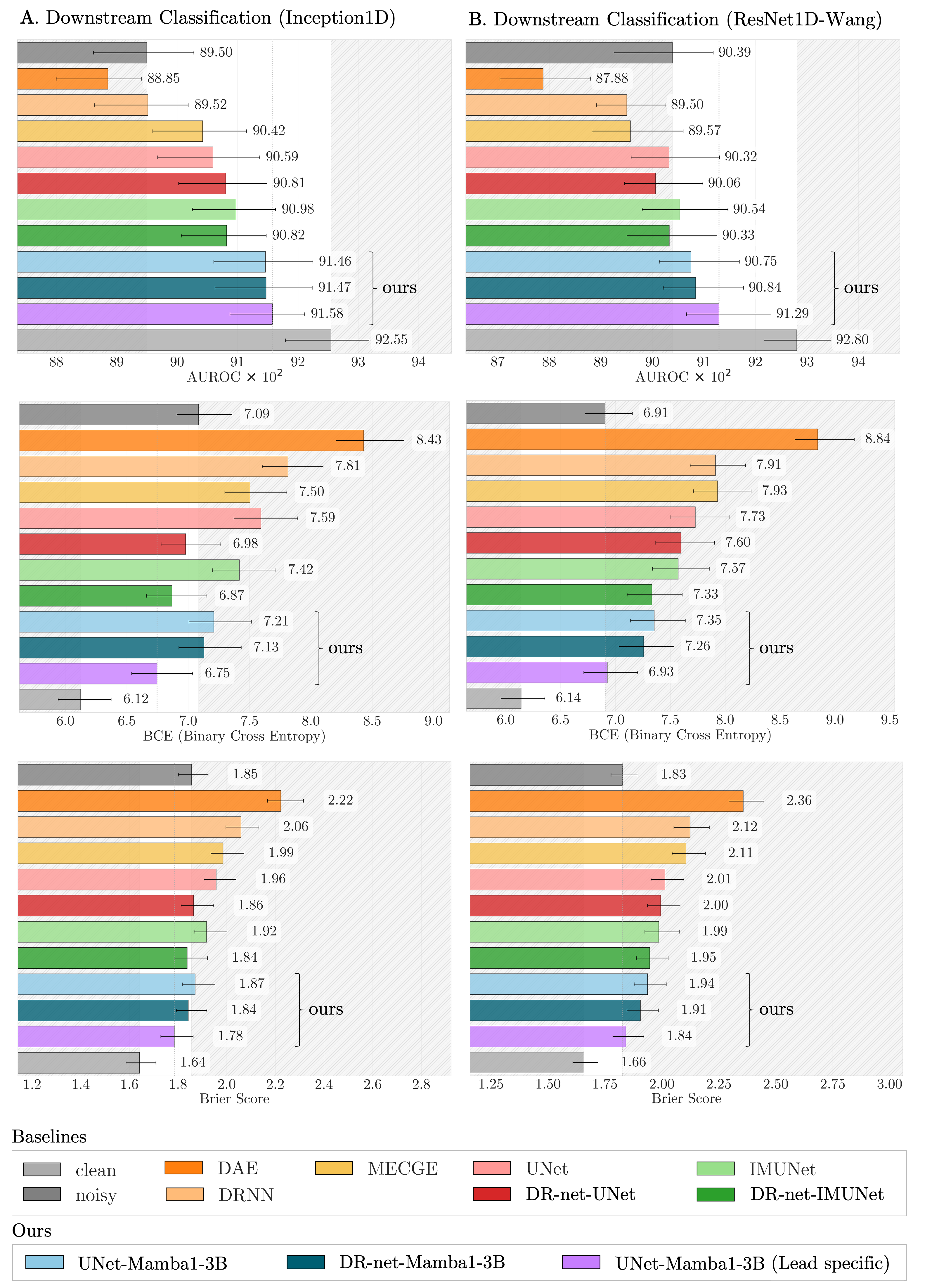}
    \caption{Downstream diagnostic classification (macro and superdiagnostic AUROC) on PTB-XL under the lower noise setting, using Inception1D and ResNet1D-Wang.}
    \label{fig:downstream_low}
\end{figure}

Table~\ref{tab:downstream_noise} reports the per-type SNR used for the lower noise setting. Figure~\ref{fig:downstream_low} shows the corresponding downstream classification results. Table~\ref{tab:per_class_default} provides the per-superdiagnostic-class AUROC breakdown, and Table~\ref{tab:macro_f1_inception_medium} reports macro sensitivity, specificity, and F1 for Inception1D. Relative model rankings are consistent with the strong noise setting.

\begin{table}[t]
\centering
\caption{Per-type SNR (dB) for the lower (medium) downstream noise setting.}
\label{tab:downstream_noise}
\small
\begin{tabular}{cccc}
\toprule
BW & MA & EM & AWGN \\
\midrule
2.5 & 7.5 & 12.5 & 22.5 \\
\bottomrule
\end{tabular}
\end{table}

\begin{table*}[t]
\centering
\caption{Per-superdiagnostic-class AUROC on PTB-XL (lower noise, Inception1D). Bold/underlined: best/second-best.}
\label{tab:per_class_default}
\small
\begin{tabular}{lccccc}
\toprule
 & CD & HYP & MI & NORM & STTC \\
\midrule
UNet Mamba1-3B & \underline{.926 $\pm$ .02} & .888 $\pm$ .02 & \underline{.916 $\pm$ .01} & .931 $\pm$ .01 & .888 $\pm$ .02 \\
DRNET Mamba1-3B & \textbf{.927 $\pm$ .02} & \textbf{.895 $\pm$ .02} & .915 $\pm$ .02 & \underline{.933 $\pm$ .01} & \underline{.889 $\pm$ .02} \\
UNet Mamba1-3B (LS) & .924 $\pm$ .01 & \underline{.894 $\pm$ .02} & \textbf{.918 $\pm$ .02} & \textbf{.935 $\pm$ .01} & \textbf{.899 $\pm$ .02} \\
\midrule
DRNN & .911 $\pm$ .01 & .887 $\pm$ .02 & .903 $\pm$ .01 & .919 $\pm$ .01 & .857 $\pm$ .02 \\
DAE & .863 $\pm$ .02 & .853 $\pm$ .02 & .886 $\pm$ .02 & .909 $\pm$ .01 & .859 $\pm$ .02 \\
UNet & .919 $\pm$ .01 & .877 $\pm$ .02 & .913 $\pm$ .01 & .926 $\pm$ .01 & .881 $\pm$ .02 \\
DRNET-UNet & .917 $\pm$ .02 & .886 $\pm$ .02 & .909 $\pm$ .02 & .927 $\pm$ .01 & .881 $\pm$ .02 \\
IMUNet & .922 $\pm$ .02 & .883 $\pm$ .02 & .910 $\pm$ .01 & .929 $\pm$ .01 & .886 $\pm$ .02 \\
DRNET-IMUNet & .919 $\pm$ .01 & .892 $\pm$ .02 & .909 $\pm$ .02 & .928 $\pm$ .01 & .887 $\pm$ .02 \\
MECGE & .916 $\pm$ .02 & .883 $\pm$ .03 & .903 $\pm$ .01 & .924 $\pm$ .01 & .888 $\pm$ .02 \\
clean & .932 $\pm$ .01 & .906 $\pm$ .02 & .925 $\pm$ .02 & .938 $\pm$ .01 & .914 $\pm$ .02 \\
noisy & .919 $\pm$ .02 & .881 $\pm$ .02 & .887 $\pm$ .02 & .923 $\pm$ .01 & .896 $\pm$ .02 \\
\bottomrule
\end{tabular}

\end{table*}

\begin{table}[hbtp]
\centering
\caption{Macro-averaged sensitivity, specificity, and F1 on PTB-XL (lower noise, Inception1D). Bold/underlined: best/second-best.}
\label{tab:macro_f1_inception_medium}
\small
\resizebox{\columnwidth}{!}{%

\begin{tabular}{lccc}
\toprule
 & \textbf{Sensitivity} & \textbf{Specificity} & \textbf{F1} \\
\midrule
UNet Mamba1-3B (ours) & .785 $\pm$ .02 & .862 $\pm$ .01 & .723 $\pm$ .01 \\
DRNET Mamba1-3B (ours) & \underline{.788 $\pm$ .02} & \textbf{.865 $\pm$ .01} & \textbf{.727 $\pm$ .01} \\
UNet Mamba1-3B (Lead aware) (ours) & \textbf{.791 $\pm$ .02} & .861 $\pm$ .01 & \underline{.725 $\pm$ .01} \\
\midrule
DAE & .698 $\pm$ .02 & .846 $\pm$ .01 & .653 $\pm$ .02 \\
DRNN & .784 $\pm$ .02 & .819 $\pm$ .01 & .682 $\pm$ .02 \\
MECGE & .760 $\pm$ .02 & .846 $\pm$ .01 & .693 $\pm$ .02 \\
UNet & .748 $\pm$ .02 & .854 $\pm$ .01 & .691 $\pm$ .02 \\
DRNET-UNet & .774 $\pm$ .02 & .859 $\pm$ .01 & .711 $\pm$ .02 \\
IMUNet & .753 $\pm$ .02 & .864 $\pm$ .01 & .703 $\pm$ .02 \\
DRNET-IMUNet & .777 $\pm$ .02 & \underline{.864 $\pm$ .01} & .717 $\pm$ .02 \\
\midrule
noisy & .857 $\pm$ .01 & .757 $\pm$ .01 & .670 $\pm$ .02 \\
clean & .815 $\pm$ .02 & .863 $\pm$ .01 & .738 $\pm$ .01 \\
\bottomrule
\end{tabular}

}
\end{table}

\subsection{Downstream Macro Results for ResNet1D-Wang} \label{app:macro_wang}

Table~\ref{tab:macro_f1_wang} reports macro-averaged sensitivity, specificity, and F1 for the ResNet1D-Wang classifier under the strong noise setting, complementing the Inception1D results in the main text.

\begin{table}[hbtp]
\centering
\caption{Macro-averaged sensitivity, specificity, and F1 on PTB-XL (strong noise, ResNet1D-Wang). Bold/underlined: best/second-best.}
\label{tab:macro_f1_wang}
\small
\resizebox{\columnwidth}{!}{%
\begin{tabular}{lccc}
\toprule
 & \textbf{Sensitivity} & \textbf{Specificity} & \textbf{F1} \\
\midrule
UNet Mamba1-3B (ours) & .797 $\pm$ .02 & .853 $\pm$ .01 & .722 $\pm$ .01 \\
DRNET Mamba1-3B (ours) & \textbf{.800 $\pm$ .02} & \textbf{.858 $\pm$ .01} & \underline{.727 $\pm$ .02} \\
UNet Mamba1-3B (Lead Specific) (ours) & \underline{.798 $\pm$ .02} & \underline{.858 $\pm$ .01} & \textbf{.728 $\pm$ .02} \\
\midrule
DAE & .733 $\pm$ .02 & .831 $\pm$ .01 & .668 $\pm$ .02 \\
DRNN & .787 $\pm$ .02 & .808 $\pm$ .01 & .679 $\pm$ .02 \\
MECGE & .786 $\pm$ .02 & .826 $\pm$ .01 & .691 $\pm$ .02 \\
UNet & .757 $\pm$ .02 & .839 $\pm$ .01 & .681 $\pm$ .02 \\
DRNET-UNet & .772 $\pm$ .02 & .841 $\pm$ .01 & .693 $\pm$ .02 \\
IMUNet & .771 $\pm$ .02 & .852 $\pm$ .01 & .707 $\pm$ .02 \\
DRNET-IMUNet & .781 $\pm$ .02 & .856 $\pm$ .01 & .713 $\pm$ .02 \\
\midrule
noisy & .831 $\pm$ .02 & .741 $\pm$ .01 & .652 $\pm$ .02 \\
clean & .808 $\pm$ .02 & .875 $\pm$ .01 & .746 $\pm$ .02 \\
\bottomrule
\end{tabular}

}
\end{table}

\subsection{Downstream Macro Results for Inception1D} \label{app:macro_inception}

\begin{figure*}[thbp]
    \centering
    \includegraphics[width=0.8\linewidth]{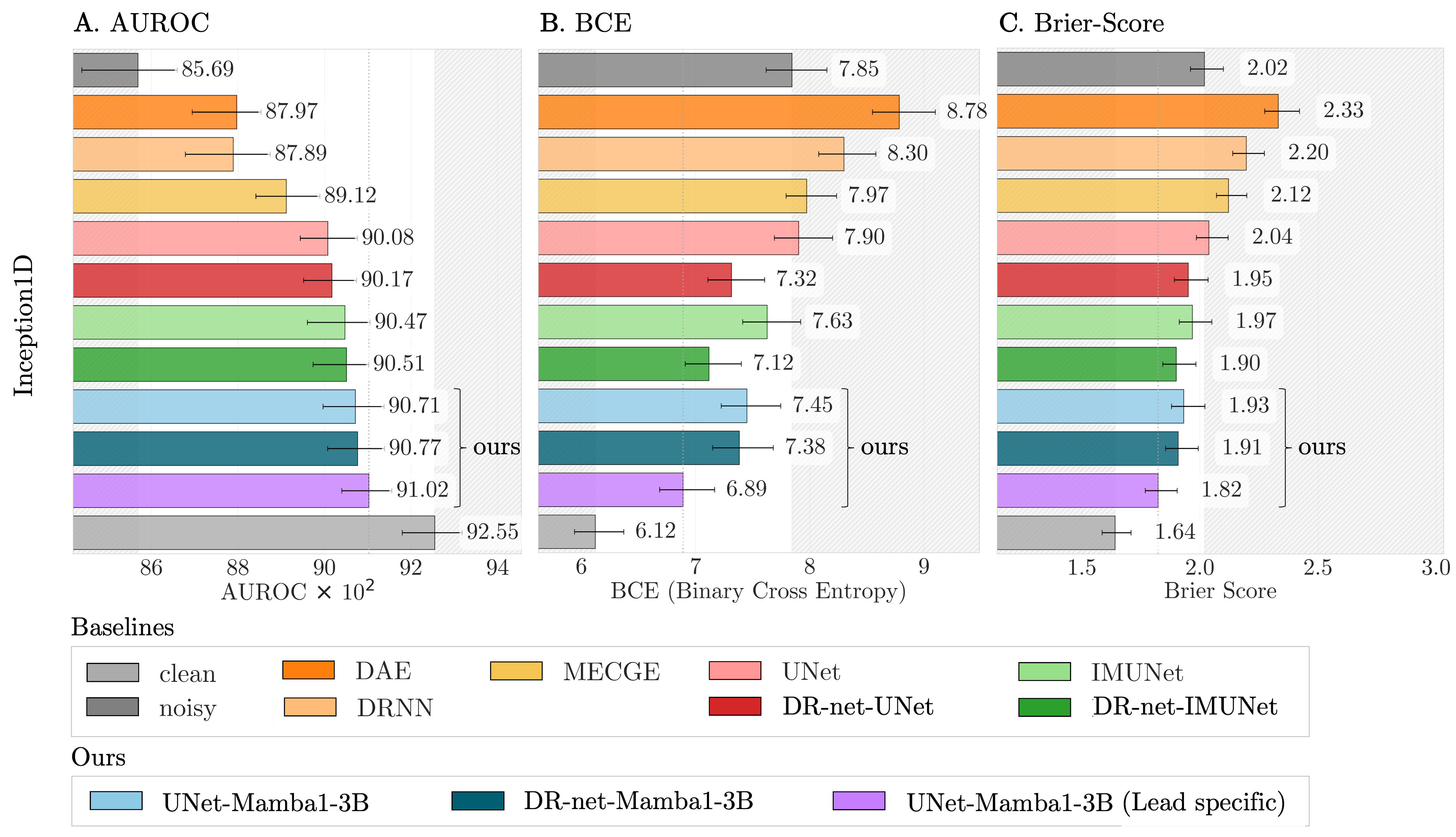}
    \caption{Downstream diagnostic classification (macro and superdiagnostic AUROC) on PTB-XL using Inception1D, comparing denoised signals against clean and noisy baselines (strong noise setting).}
    \label{fig:downstream_high_inception1D}
\end{figure*}

Figure~\ref{fig:downstream_high_inception1D} summarizes downstream classification performance on PTB-XL under the strong noise setting (BW 0$dB$; MA 5$dB$; EM 10$dB$; AWGN 20$dB$) using the Inception1D backbone.

\subsection{Superdiagnostic Sensitivity, Specificity, and F1} \label{app:superdiagnostic_spec}

Tables~\ref{tab:per_class_f1_inception} and~\ref{tab:per_class_f1_wang} report per-superdiagnostic-class sensitivity, specificity, and F1 under the strong noise setting for Inception1D and ResNet1D-Wang, respectively.

\begin{table}[hbtp]
\centering
\caption{Per-superdiagnostic-class sensitivity, specificity, and F1 on PTB-XL (strong noise, Inception1D). Bold/underlined: best/second-best.}
\label{tab:per_class_f1_inception}
\small
\resizebox{\columnwidth}{!}{%
\begin{tabular}{lccccc}
\toprule
 & CD & HYP & MI & NORM & STTC \\
\midrule
\multicolumn{6}{l}{\textit{Sensitivity}} \\
\midrule
UNet-Mamba1-3B  & \underline{.822 $\pm$ .03} & .526 $\pm$ .05 & .865 $\pm$ .03 & .918 $\pm$ .02 & .792 $\pm$ .03 \\
DR-net-Mamba1-3B  & \textbf{.834 $\pm$ .04} & \underline{.534 $\pm$ .05} & .854 $\pm$ .04 & .917 $\pm$ .02 & .801 $\pm$ .04 \\
UNet-Mamba1-3B (LS)  & .816 $\pm$ .03 & \textbf{.556 $\pm$ .05} & .815 $\pm$ .03 & \textbf{.943 $\pm$ .01} & .824 $\pm$ .04 \\
\midrule
DAE & .640 $\pm$ .05 & .299 $\pm$ .07 & \textbf{.889 $\pm$ .03} & .891 $\pm$ .02 & .769 $\pm$ .04 \\
DRNN & .822 $\pm$ .03 & .489 $\pm$ .06 & \underline{.887 $\pm$ .03} & .863 $\pm$ .03 & \textbf{.860 $\pm$ .03} \\
MECGE & .796 $\pm$ .03 & .440 $\pm$ .05 & .850 $\pm$ .03 & .920 $\pm$ .02 & .792 $\pm$ .04 \\
UNet & .778 $\pm$ .03 & .351 $\pm$ .06 & .872 $\pm$ .03 & \underline{.920 $\pm$ .02} & .818 $\pm$ .04 \\
DR-net-UNet & .816 $\pm$ .03 & .496 $\pm$ .06 & .835 $\pm$ .04 & .887 $\pm$ .02 & \underline{.833 $\pm$ .03} \\
IMUNet & .794 $\pm$ .03 & .407 $\pm$ .05 & .874 $\pm$ .03 & .916 $\pm$ .02 & .773 $\pm$ .04 \\
DR-net-IMUNet & .814 $\pm$ .03 & .526 $\pm$ .05 & .841 $\pm$ .04 & .895 $\pm$ .03 & .809 $\pm$ .04 \\
\midrule
noisy & .871 $\pm$ .03 & .642 $\pm$ .06 & .891 $\pm$ .02 & .939 $\pm$ .02 & .941 $\pm$ .02 \\
clean & .846 $\pm$ .03 & .627 $\pm$ .05 & .807 $\pm$ .04 & .946 $\pm$ .01 & .847 $\pm$ .03 \\
\midrule
\multicolumn{6}{l}{\textit{Specificity}} \\
\midrule
UNet-Mamba1-3B  & .902 $\pm$ .01 & .964 $\pm$ .01 & .803 $\pm$ .02 & .783 $\pm$ .03 & \underline{.860 $\pm$ .02} \\
DR-net-Mamba1-3B  & .905 $\pm$ .01 & .963 $\pm$ .01 & .819 $\pm$ .02 & .785 $\pm$ .03 & .851 $\pm$ .02 \\
UNet-Mamba1-3B (LS)  & .899 $\pm$ .01 & .956 $\pm$ .01 & \textbf{.847 $\pm$ .02} & .763 $\pm$ .03 & .840 $\pm$ .02 \\
\midrule
DAE & \textbf{.945 $\pm$ .01} & \textbf{.990 $\pm$ .00} & .728 $\pm$ .02 & .757 $\pm$ .03 & .809 $\pm$ .02 \\
DRNN & .873 $\pm$ .01 & .969 $\pm$ .01 & .730 $\pm$ .03 & \textbf{.825 $\pm$ .02} & .697 $\pm$ .02 \\
MECGE & .889 $\pm$ .02 & .972 $\pm$ .01 & .778 $\pm$ .02 & .767 $\pm$ .03 & .824 $\pm$ .02 \\
UNet & \underline{.927 $\pm$ .01} & \underline{.984 $\pm$ .01} & .796 $\pm$ .02 & .767 $\pm$ .02 & .794 $\pm$ .02 \\
DR-net-UNet & .901 $\pm$ .01 & .968 $\pm$ .01 & \underline{.826 $\pm$ .02} & .809 $\pm$ .02 & .789 $\pm$ .02 \\
IMUNet & .916 $\pm$ .01 & .972 $\pm$ .01 & .790 $\pm$ .02 & .774 $\pm$ .02 & \textbf{.869 $\pm$ .02} \\
DR-net-IMUNet & .898 $\pm$ .02 & .965 $\pm$ .01 & .814 $\pm$ .02 & \underline{.815 $\pm$ .02} & .829 $\pm$ .02 \\
\midrule
noisy & .819 $\pm$ .02 & .915 $\pm$ .02 & .687 $\pm$ .02 & .724 $\pm$ .03 & .642 $\pm$ .02 \\
clean & .887 $\pm$ .02 & .936 $\pm$ .01 & .881 $\pm$ .01 & .765 $\pm$ .03 & .846 $\pm$ .02 \\
\midrule
\multicolumn{6}{l}{\textit{F1}} \\
\midrule
UNet-Mamba1-3B  & .765 $\pm$ .03 & .591 $\pm$ .04 & .705 $\pm$ .03 & .839 $\pm$ .02 & \textbf{.713 $\pm$ .03} \\
DR-net-Mamba1-3B  & \textbf{.776 $\pm$ .03} & \underline{.596 $\pm$ .04} & \underline{.714 $\pm$ .03} & .839 $\pm$ .02 & .710 $\pm$ .03 \\
UNet-Mamba1-3B (LS)  & .758 $\pm$ .03 & \textbf{.596 $\pm$ .05} & \textbf{.718 $\pm$ .03} & \textbf{.843 $\pm$ .02} & \underline{.712 $\pm$ .04} \\
\midrule
DAE & .703 $\pm$ .04 & .436 $\pm$ .07 & .659 $\pm$ .03 & .812 $\pm$ .02 & .653 $\pm$ .03 \\
DRNN & .733 $\pm$ .03 & .572 $\pm$ .05 & .659 $\pm$ .03 & .829 $\pm$ .02 & .617 $\pm$ .03 \\
MECGE & .735 $\pm$ .03 & .539 $\pm$ .05 & .677 $\pm$ .03 & .833 $\pm$ .02 & .680 $\pm$ .03 \\
UNet & \underline{.770 $\pm$ .03} & .480 $\pm$ .06 & .704 $\pm$ .03 & .832 $\pm$ .02 & .668 $\pm$ .03 \\
DR-net-UNet & .761 $\pm$ .03 & .577 $\pm$ .05 & .710 $\pm$ .03 & .835 $\pm$ .02 & .672 $\pm$ .03 \\
IMUNet & .765 $\pm$ .03 & .508 $\pm$ .05 & .699 $\pm$ .03 & .834 $\pm$ .02 & .711 $\pm$ .04 \\
DR-net-IMUNet & .755 $\pm$ .03 & .594 $\pm$ .05 & .703 $\pm$ .03 & \underline{.842 $\pm$ .02} & .693 $\pm$ .03 \\
\midrule
noisy & .704 $\pm$ .04 & .574 $\pm$ .05 & .631 $\pm$ .03 & .823 $\pm$ .02 & .619 $\pm$ .03 \\
clean & .761 $\pm$ .03 & .604 $\pm$ .05 & .747 $\pm$ .03 & .845 $\pm$ .02 & .730 $\pm$ .03 \\
\bottomrule
\end{tabular}

}
\end{table}

\begin{table}[t]
\centering
\caption{Per-superdiagnostic-class sensitivity, specificity, and F1 on PTB-XL (strong noise, ResNet1D-Wang). Bold/underlined: best/second-best.}
\label{tab:per_class_f1_wang}
\small
\resizebox{\columnwidth}{!}{%
\begin{tabular}{lccccc}
\toprule
 & CD & HYP & MI & NORM & STTC \\
\midrule
\multicolumn{6}{l}{\textit{Sensitivity}} \\
\midrule
UNet-Mamba1-3B  & \textbf{.784 $\pm$ .03} & .515 $\pm$ .06 & .898 $\pm$ .02 & \underline{.942 $\pm$ .02} & .786 $\pm$ .03 \\
DR-net-Mamba1-3B  & \underline{.784 $\pm$ .03} & \underline{.522 $\pm$ .07} & .870 $\pm$ .03 & .931 $\pm$ .02 & .794 $\pm$ .04 \\
UNet-Mamba1-3B (LS)  & .780 $\pm$ .03 & \textbf{.541 $\pm$ .06} & .841 $\pm$ .03 & \textbf{.961 $\pm$ .01} & .803 $\pm$ .04 \\
\midrule
DAE & .667 $\pm$ .04 & .347 $\pm$ .06 & .894 $\pm$ .03 & .915 $\pm$ .02 & .756 $\pm$ .04 \\
DRNN & .762 $\pm$ .03 & .507 $\pm$ .06 & .902 $\pm$ .02 & .898 $\pm$ .03 & \textbf{.850 $\pm$ .03} \\
MECGE & .774 $\pm$ .03 & .440 $\pm$ .05 & .896 $\pm$ .03 & .942 $\pm$ .02 & .780 $\pm$ .04 \\
UNet & .760 $\pm$ .03 & .362 $\pm$ .06 & \textbf{.906 $\pm$ .03} & .931 $\pm$ .02 & \underline{.822 $\pm$ .03} \\
DR-net-UNet & .766 $\pm$ .04 & .451 $\pm$ .06 & .887 $\pm$ .04 & .907 $\pm$ .02 & .807 $\pm$ .04 \\
IMUNet & .743 $\pm$ .03 & .429 $\pm$ .06 & \underline{.904 $\pm$ .02} & .932 $\pm$ .02 & .771 $\pm$ .04 \\
DR-net-IMUNet & .764 $\pm$ .03 & .478 $\pm$ .06 & .874 $\pm$ .04 & .929 $\pm$ .02 & .786 $\pm$ .04 \\
\midrule
noisy & .808 $\pm$ .03 & .627 $\pm$ .05 & .896 $\pm$ .03 & .939 $\pm$ .01 & .915 $\pm$ .03 \\
clean & .820 $\pm$ .03 & .604 $\pm$ .06 & .807 $\pm$ .04 & .959 $\pm$ .02 & .848 $\pm$ .04 \\
\midrule
\multicolumn{6}{l}{\textit{Specificity}} \\
\midrule
UNet-Mamba1-3B  & .924 $\pm$ .01 & .968 $\pm$ .01 & .794 $\pm$ .02 & .788 $\pm$ .02 & \underline{.885 $\pm$ .02} \\
DR-net-Mamba1-3B  & .924 $\pm$ .01 & .964 $\pm$ .01 & \underline{.812 $\pm$ .02} & .793 $\pm$ .03 & .875 $\pm$ .02 \\
UNet-Mamba1-3B (LS)  & .925 $\pm$ .01 & .960 $\pm$ .01 & \textbf{.845 $\pm$ .02} & .767 $\pm$ .03 & .874 $\pm$ .02 \\
\midrule
DAE & \textbf{.938 $\pm$ .01} & \underline{.988 $\pm$ .00} & .699 $\pm$ .02 & .766 $\pm$ .03 & .855 $\pm$ .02 \\
DRNN & .925 $\pm$ .01 & .970 $\pm$ .01 & .751 $\pm$ .02 & \underline{.824 $\pm$ .02} & .756 $\pm$ .03 \\
MECGE & .905 $\pm$ .01 & .974 $\pm$ .01 & .757 $\pm$ .02 & .768 $\pm$ .02 & .869 $\pm$ .02 \\
UNet & \underline{.935 $\pm$ .01} & \textbf{.989 $\pm$ .01} & .766 $\pm$ .02 & .788 $\pm$ .02 & .840 $\pm$ .02 \\
DR-net-UNet & .932 $\pm$ .01 & .972 $\pm$ .01 & .770 $\pm$ .02 & \textbf{.825 $\pm$ .02} & .835 $\pm$ .02 \\
IMUNet & .930 $\pm$ .01 & .978 $\pm$ .01 & .763 $\pm$ .02 & .800 $\pm$ .02 & \textbf{.890 $\pm$ .01} \\
DR-net-IMUNet & .921 $\pm$ .01 & .971 $\pm$ .01 & .792 $\pm$ .02 & .821 $\pm$ .02 & .872 $\pm$ .02 \\
\midrule
noisy & .879 $\pm$ .02 & .934 $\pm$ .01 & .693 $\pm$ .02 & .767 $\pm$ .02 & .693 $\pm$ .02 \\
clean & .911 $\pm$ .01 & .943 $\pm$ .01 & .881 $\pm$ .01 & .776 $\pm$ .02 & .865 $\pm$ .02 \\
\midrule
\multicolumn{6}{l}{\textit{F1}} \\
\midrule
UNet-Mamba1-3B  & \underline{.769 $\pm$ .03} & .591 $\pm$ .06 & .715 $\pm$ .03 & .854 $\pm$ .02 & \textbf{.735 $\pm$ .03} \\
DR-net-Mamba1-3B  & \textbf{.770 $\pm$ .03} & .588 $\pm$ .06 & \underline{.716 $\pm$ .03} & .850 $\pm$ .02 & .729 $\pm$ .03 \\
UNet-Mamba1-3B (LS)  & .769 $\pm$ .03 & \textbf{.593 $\pm$ .05} & \textbf{.730 $\pm$ .03} & .854 $\pm$ .02 & \underline{.734 $\pm$ .03} \\
\midrule
DAE & .712 $\pm$ .03 & .486 $\pm$ .07 & .641 $\pm$ .03 & .829 $\pm$ .02 & .687 $\pm$ .03 \\
DRNN & .758 $\pm$ .03 & \underline{.591 $\pm$ .05} & .683 $\pm$ .03 & .848 $\pm$ .02 & .655 $\pm$ .03 \\
MECGE & .740 $\pm$ .04 & .544 $\pm$ .05 & .684 $\pm$ .03 & .845 $\pm$ .02 & .715 $\pm$ .04 \\
UNet & .768 $\pm$ .04 & .503 $\pm$ .05 & .696 $\pm$ .03 & .848 $\pm$ .02 & .711 $\pm$ .03 \\
DR-net-UNet & .769 $\pm$ .03 & .548 $\pm$ .06 & .690 $\pm$ .03 & .853 $\pm$ .02 & .698 $\pm$ .04 \\
IMUNet & .753 $\pm$ .03 & .542 $\pm$ .05 & .693 $\pm$ .03 & \underline{.855 $\pm$ .01} & .731 $\pm$ .04 \\
DR-net-IMUNet & .754 $\pm$ .03 & .568 $\pm$ .06 & .701 $\pm$ .03 & \textbf{.863 $\pm$ .02} & .722 $\pm$ .03 \\
\midrule
noisy & .731 $\pm$ .03 & .600 $\pm$ .05 & .638 $\pm$ .03 & .843 $\pm$ .02 & .641 $\pm$ .04 \\
clean & .775 $\pm$ .03 & .603 $\pm$ .05 & .747 $\pm$ .03 & .857 $\pm$ .02 & .750 $\pm$ .04 \\
\bottomrule
\end{tabular}

}
\end{table}

\subsection{Per-Class Downstream Classification (ResNet1D-Wang)} \label{app:downstream_wang}

\begin{figure*}[tbhp]
    \centering
    \includegraphics[width=\linewidth]{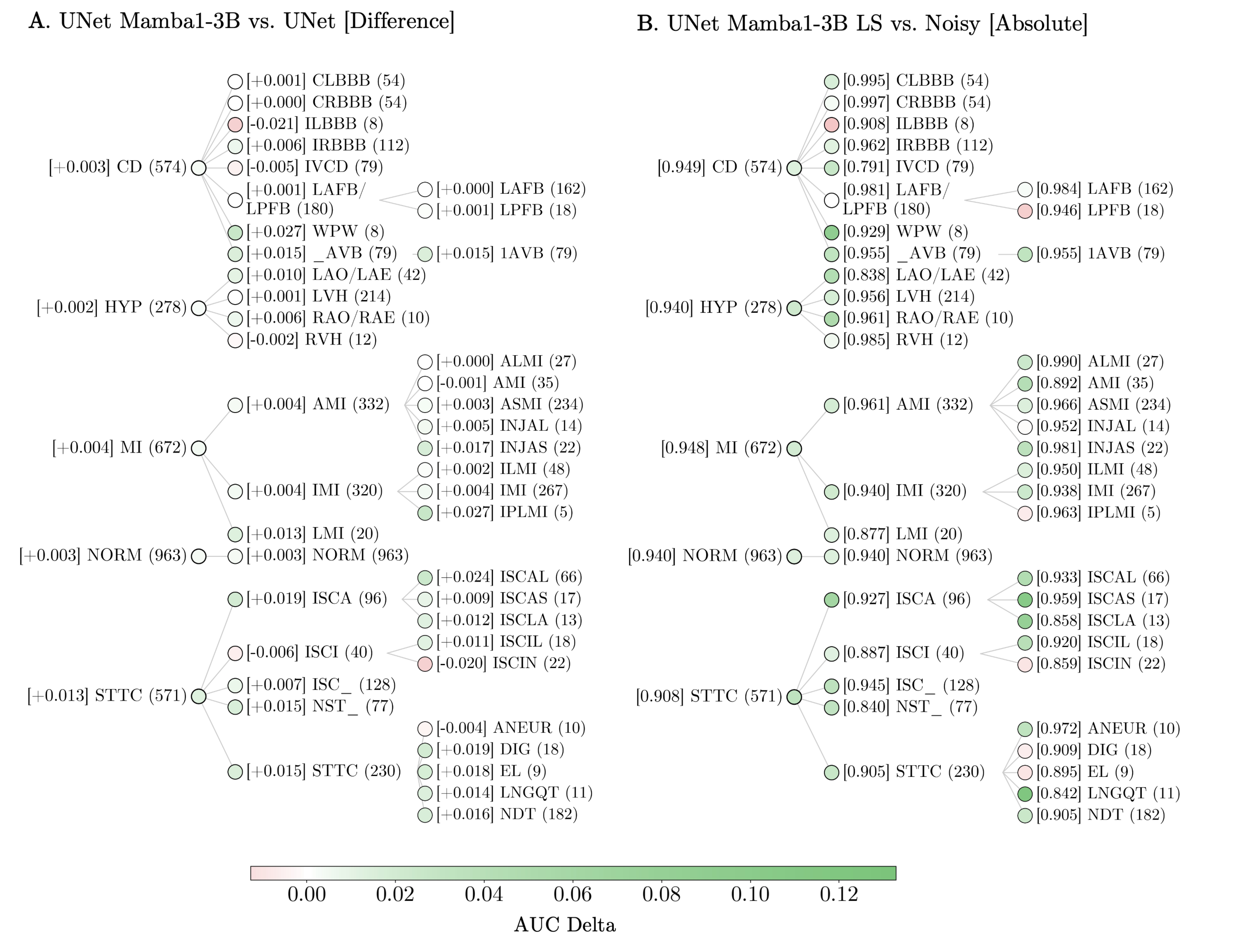}
    \caption{\textbf{Per-diagnostic-class downstream AUROC (ResNet1D-Wang).} \textbf{(A)} AUROC delta when adding a Mamba bottleneck to the UNet, all other design choices held constant. \textbf{(B)} Absolute AUROC of the lead-specific UNet-Mamba; color indicates improvement over the noisy baseline. In both panels, diagnostic classes with fewer than five segments are omitted. See Table~\ref{tab:form_labels} for abbreviations.}
    \label{fig:tree_wang}
\end{figure*}

Figure~\ref{fig:tree_wang} shows the per-diagnostic-class AUROC breakdown using ResNet1D-Wang, analogous to Figure~\ref{fig:tree_inception} in the main text. Panel~(A) shows the AUROC delta when adding a Mamba bottleneck; panel~(B) shows absolute AUROC of the lead-specific UNet-Mamba.

\subsection{Diagnostic Abbreviations} \label{app:abbreviations}
The diagnostic class abbreviations used in Figures~\ref{fig:tree_inception} and~\ref{fig:tree_wang} are listed in Table~\ref{tab:form_labels}.
\begin{table*}[t]
\centering
\caption{ECG Form Statement Labels}
\label{tab:form_labels}
\scriptsize
\setlength{\tabcolsep}{3pt}
\renewcommand{\arraystretch}{0.92}

\begin{tabularx}{\columnwidth}{@{}lX@{}}
\toprule
Abbreviation & Description \\
\midrule
NDT & non-diagnostic T abnormalities \\
NST\_ & non-specific ST changes \\
DIG & digitalis-effect \\
LNGQT & long QT-interval \\
NORM & normal ECG \\
IMI & inferior myocardial infarction \\
ASMI & anteroseptal myocardial infarction \\
LVH & left ventricular hypertrophy \\
LAFB & left anterior fascicular block \\
ISC\_ & non-specific ischemic \\
IRBBB & incomplete right bundle branch block \\
1AVB & first degree AV block \\
IVCD & non-specific intraventricular conduction disturbance (block) \\
ISCAL & ischemic in anterolateral leads \\
CRBBB & complete right bundle branch block \\
CLBBB & complete left bundle branch block \\
ILMI & inferolateral myocardial infarction \\
LAO/LAE & left atrial overload/enlargement \\
AMI & anterior myocardial infarction \\
ALMI & anterolateral myocardial infarction \\
ISCIN & ischemic in inferior leads \\
INJAS & subendocardial injury in anteroseptal leads \\
LMI & lateral myocardial infarction \\
ISCIL & ischemic in inferolateral leads \\
LPFB & left posterior fascicular block \\
ISCAS & ischemic in anteroseptal leads \\
INJAL & subendocardial injury in anterolateral leads \\
ISCLA & ischemic in lateral leads \\
RVH & right ventricular hypertrophy \\
ANEUR & ST-T changes compatible with ventricular aneurysm \\
RAO/RAE & right atrial overload/enlargement \\
EL & electrolytic disturbance or drug (former EDIS) \\
WPW & Wolff-Parkinson-White syndrome \\
ILBBB & incomplete left bundle branch block \\
IPLMI & inferoposterolateral myocardial infarction \\
ISCAN & ischemic in anterior leads \\
IPMI & inferoposterior myocardial infarction \\
SEHYP & septal hypertrophy \\
INJIN & subendocardial injury in inferior leads \\
INJLA & subendocardial injury in lateral leads \\
PMI & posterior myocardial infarction \\
3AVB & third degree AV block \\
INJIL & subendocardial injury in inferolateral leads \\
2AVB & second degree AV block \\
\bottomrule
\end{tabularx}
\end{table*}

\subsection{STTC and CD diagnostic morphologies and The Mamba Bottleneck}

Table \ref{tab:sttc_vs_cd} shows a comparison of STTC and CD diagnostic morphologies and their interaction with the Mamba bottleneck.

\begin{table*}[thbp]
\centering
\caption{Comparison of STTC and CD diagnostic morphologies and their interaction with the Mamba bottleneck.}
\label{tab:sttc_vs_cd}
\small

\begin{tabular}{lcc}
\toprule
 & STTC & CD \\
\midrule
Temporal scale & Broad (200--400\,ms) & Narrow (R-peak time 45--60\,ms) \\
Bottleneck steps spanned & ${\sim}4$--7 & ${\sim}1$ \\
Frequency content & Low & High \\
Context needed & Global (baseline-relative) & Local (QRS-internal) \\
Mamba effect on denoising & Better noise/signal separation & Over-smoothing of sharp features \\
\bottomrule
\end{tabular}
\end{table*}

\end{document}